%% file: main.tex
\documentclass[10pt,twocolumn,letterpaper]{article}

\usepackage{iccv}              


\definecolor{iccvblue}{rgb}{0.21,0.49,0.74}
\usepackage[pagebackref,breaklinks,colorlinks,allcolors=iccvblue]{hyperref}

\usepackage{graphicx}
\usepackage{booktabs}
\usepackage{algorithm}  
\usepackage{algorithmic}
\usepackage{multicol}
\usepackage{multirow}
\usepackage[normalem]{ulem}
\useunder{\uline}{\ul}{}
\usepackage{amsmath}
\usepackage{wrapfig}
\usepackage{bbm}

\newcommand{\mcX}{\mathcal{X}}
\newcommand{\mcY}{\mathcal{Y}}

\newcommand{\OURS}{ODP-Bench}
\newcommand{\NUMDATA}{29}
\newcommand{\NUMALG}{10}
\newcommand{\NUMMODEL}{1,444}

\newcommand\blfootnote[1]{
    \begingroup
    \renewcommand\thefootnote{}\footnote{#1}
    \addtocounter{footnote}{-1}
    \endgroup
}

\def\paperID{2691} 
\def\confName{ICCV}
\def\confYear{2025}

\title{\OURS: Benchmarking Out-of-Distribution Performance Prediction}

\author{Han Yu$^{1,\dag}$, Kehan Li$^{1,\dag}$, Dongbai Li$^1$, Yue He$^2$, Xingxuan Zhang$^1$, Peng Cui$^{1,*}$\\
$^1$Department of Computer Science, Tsinghua University\\
$^2$School of Information, Renmin University of China\\
{\tt\small yuh21@mails.tsinghua.edu.cn, lkh20@mails.tsinghua.edu.cn, ldb22@mails.tsinghua.edu.cn}\\
{\tt\small hy865865@gmail.com, xingxuanzhang@hotmail.com, cuip@tsinghua.edu.cn}
}

\begin{document}
\maketitle

\input{sec/0_abstract}    

\blfootnote{$\dag$Equal contribution, *Corresponding author}

\input{sec/1_intro}

\input{sec/2_related}

\input{sec/3_1_pre}

\input{sec/3_2_method}

\input{sec/4_exp}

\input{sec/5_con}

\section*{Acknowledgement}

This work was supported by NSFC (No. 62425206), Tsinghua-Toyota Joint Research Fund, NSFC (No. 62141607), and Beijing Municipal Science and Technology Project (No. Z241100004224009).
Peng Cui is the corresponding author. All opinions in this paper are those of the authors and do not necessarily reflect the views of the funding agencies.

\appendix

\renewcommand{\twocolumn}[1][]{\onecolumn}  

\input{sec/X_suppl}

{
    \small
    \bibliographystyle{ieeenat_fullname}
    \bibliography{main}
}

\end{document}

%% file: sec/0_abstract.tex
\begin{abstract}

Recently, there has been gradually more attention paid to Out-of-Distribution (OOD) performance prediction, whose goal is to predict the performance of trained models on unlabeled OOD test datasets, so that we could better leverage and deploy off-the-shelf trained models in risk-sensitive scenarios. 
Although progress has been made in this area, evaluation protocols in previous literature are inconsistent, and most works cover only a limited number of real-world OOD datasets and types of distribution shifts. 
To provide convenient and fair comparisons for various algorithms, we propose Out-of-Distribution Performance Prediction Benchmark (\OURS), a comprehensive benchmark that includes most commonly used OOD datasets and existing practical performance prediction algorithms. 
We provide our trained models as a testbench for future researchers, thus guaranteeing the consistency of comparison and avoiding the burden of repeating the model training process. 
Furthermore, we also conduct in-depth experimental analyses to better understand their capability boundary. 

\end{abstract}

%% file: sec/1_intro.tex
\section{Introduction}
\label{sec:intro}

Although deep learning has achieved significant progress in many applications~\citep{brown2020language,achiam2023gpt, liu2023visual, tian2025yolov12}, their performance heavily relies on the assumption that test data follows the same distribution as training data does, known as the I.I.D. assumption. Yet in wild environments, such an assumption can be easily violated and models are likely to encounter severe performance degradation in the face of distribution shifts~\citep{shen2021towards}, which severely hinders applications of deep learning models in risk-sensitive areas like autonomous driving~\citep{chen2024end} and medical imaging~\citep{yang2024limits}. 
In recent years, there have been many algorithms proposed to improve the Out-of-Distribution (OOD) generalization ability of models, including invariant learning~\citep{arjovsky2019invariant,krueger2021out,creager2021environment}, domain generalization~\citep{cha2021swad,rame2022fishr,huang2020self}, distributionally robust optimization~\citep{duchi2021learning,volpi2018generalizing,sinha2018certifying}, stable learning~\citep{zhang2021deep,yu2023stable, cui2022stable}, etc. 
Nevertheless, none of these algorithms could substantially improve the OOD performance~\citep{gulrajani2020search,idrissi2022simple} and learn trustworthy models that can be deployed under risk-sensitive scenarios.

\begin{figure}[t]
	\centering
        \includegraphics[width=0.8\linewidth]{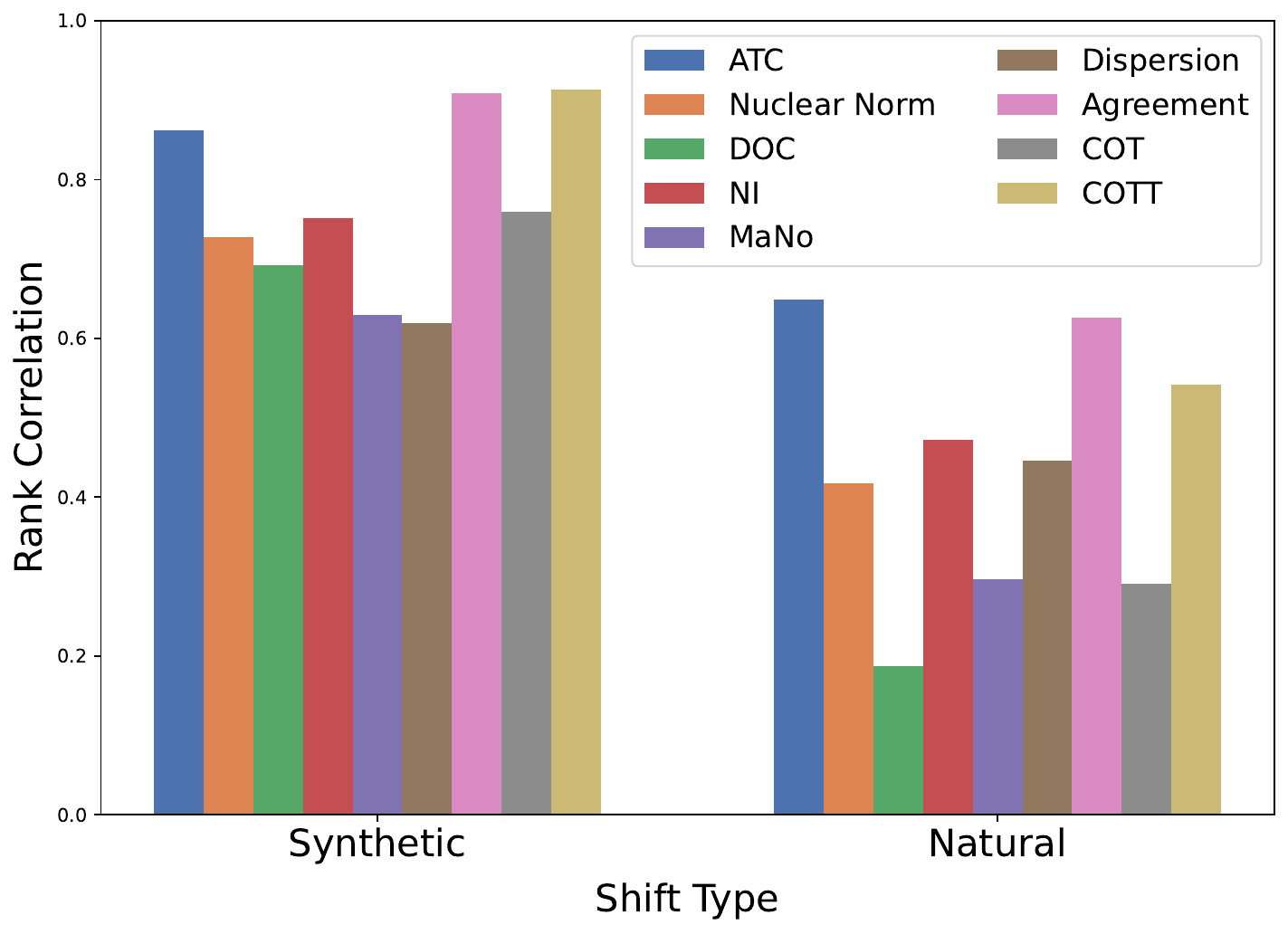}
	\caption{Rank correlation between predicted scores and ground truth performances. 
    We can see that most of current algorithms do well in synthetic corruptions, but generally fail in natural shifts.}
	\label{fig:shift}
\end{figure}

Instead of developing algorithms and training new models to improve the OOD generalization ability, with a tremendous number of off-the-shelf trained models nowadays, it is efficient and meaningful to directly leverage them by taking a look at the other side of the coin~\citep{yu2024survey}: Given trained models, we aim to predict their performance on unlabeled OOD datasets, i.e. \textit{OOD performance prediction}. 
With successful performance prediction, we could safely apply models to their well-performing scenarios and use them with caution in their poorly-performing scenarios, e.g. simply avoiding their usage or cooperating with humans in such scenarios. Meanwhile, we could conduct model selection from a pool of models when facing unseen environments. Therefore, we could broaden the application of off-the-shelf deep learning models in risk-sensitive areas.

Recently, more attention has been paid to the area of OOD performance prediction from various perspectives. They try to leverage model confidence~\citep{guillory2021predicting,garg2022leveraging}, distribution discrepancy~\citep{deng2021labels,lu2023characterizing}, model agreement~\citep{baek2022agreement,rosenfeld2023almost}, etc., to carry out the task of performance prediction. Some literature tries to predict the value of model performance~\citep{deng2021does,garg2022leveraging,ng2024predicting} while more proposes to calculate a score as the surrogate of performance~\citep{xie2023importance,peng2024energy,xie2024mano}. 
Although their improvement and effectiveness have been preliminarily verified through experiments on real-world datasets~\citep{hendrycks2019benchmarking,recht2018cifar,recht2019imagenet,hendrycks2021natural}, the evaluation protocols in previous works are inconsistent, including details of training models whose performance needs to be predicted, OOD test datasets, and evaluation metrics. 
Meanwhile, both the number of real-world test datasets and the types of distribution shifts are relatively limited. 
For example, datasets widely used in domain generalization~\citep{li2017deeper,khosla2012undoing,venkateswara2017deep} and subpopulation shift~\citep{liu2015deep,irvin2019chexpert,sagawa2020distributionally}, which are two vital sub-areas in OOD generalization, are rarely covered in previous works of OOD performance prediction. 
In such cases, it becomes difficult to make fair and comprehensive comparisons between different performance prediction algorithms. 
Moreover, the application range and capability boundary of current performance prediction algorithms have not been well-determined and clearly explored. 

To fully understand and investigate existing algorithms and better promote future research in this area, in this paper we establish Out-of-Distribution Performance Prediction Benchmark (\OURS). 
We provide \NUMMODEL~trained models of various architectures, weight initializations, and training algorithms, which are open-sourced and directly available for future researchers so that they do not need to repeat the model training process. 
Our benchmark includes \NUMDATA~OOD datasets that cover diverse types of distribution shifts, and \NUMALG~performance prediction algorithms.
Besides, we provide our codebase\footnote{\hyperlink{https://github.com/h-yu16/Performance_Prediction/}{https://github.com/h-yu16/Performance\_Prediction/}} where a new proposed algorithm can be easily added with small efforts of code implementation. 
In this way, we enable convenient and fair comparisons for both current and future algorithms. 
Furthermore, we conduct empirical analyses of performance prediction algorithms. 
One pivotal conclusion is that, as revealed in~\Cref{fig:shift}, current algorithms show their effectiveness on OOD datasets with synthetic corruptions, which are exactly the most commonly adopted datasets in previous works of performance prediction, but cannot address diverse and complex natural distribution shifts well. 
This suggests the necessity and significance of our comprehensive benchmark for promoting future research on performance prediction. 
Our contributions are listed below:
\begin{itemize}
    \item We propose \OURS, a large comprehensive benchmark of OOD performance prediction covering \NUMDATA~OOD datasets and \NUMALG~algorithms, where diverse types of distribution shifts are included. 
    \item Our benchmark provides \NUMMODEL~off-the-shelf models and a codebase for future researchers to easily add and test newly proposed algorithms on. This enables fair and convenient comparisons between different algorithms.
    \item We conduct further experimental analyses to better understand the capability boundary of current algorithms. 
\end{itemize}

%% file: sec/2_related.tex
\section{Related Work}
\label{sec:related}

\paragraph{Out-of-Distribution Generalization} There have been multiple branches of research devoted to OOD generalization of machine learning models~\citep{shen2021towards}. 
Invariant learning aims to capture the invariant relationship between the outcome and input covariates given training data from different environments~\citep{arjovsky2019invariant,krueger2021out,koyama2020out}, but usually limited in the range of tabular data. 
Distributionally robust optimization (DRO) pursues a different objective, i.e. finding the worst performing distribution in a ball centered around the training distribution and optimizing under it~\citep{duchi2021learning, sinha2018certifying}, yet it suffers from over-pessimism.
Focusing on visual data, domain generalization proposes diverse solutions, including invariant representation learning~\citep{shankar2018generalizing,rame2022fishr}, data augmentation~\citep{zhou2020domain}, meta learning~\citep{li2018learning,balaji2018metareg,li2019episodic}, flatness-aware optimization~\citep{cha2021swad,zhang2023gradient,zhang2023flatness}, etc. 
Another branch is stable learning which aims to decorrelate covariates via sample reweighting to remove spurious correlations~\citep{zhang2021deep,cui2022stable} but suffers from variance inflation~\citep{yu2023stable, yu2025sample}.

\paragraph{Out-of-Distribution Performance Prediction} Despite progress has been made in OOD generalization, none of current algorithms have empirically shown a significant improvement over simple empirical risk minimization (ERM)~\citep{gulrajani2020search, idrissi2022simple}. As suggested~\citep{yu2024survey}, we can put more emphasis on the evaluation of OOD generalization, among which OOD performance prediction is an important aspect. 
Some take advantage of different model output properties like model confidence~\citep{garg2022leveraging}, prediction dispersity~\citep{deng2023confidence}, matrix norm~\citep{xie2024mano}, feature separability~\citep{xie2023importance}, neighborhood invariance~\citep{ng2024predicting}, rotation invariance~\citep{deng2021does}, etc. Some measures the discrepancy between covariates of training distribution and those of OOD test distribution~\citep{deng2021labels,lu2023characterizing}. Another series leverages the phenomenon of agreement-on-the-line~\citep{baek2022agreement, rosenfeld2023almost} given a number of trained models.

%% file: sec/3_1_pre.tex
\begin{table*}[htbp]
\centering
\caption{Information of all datasets included in \OURS. For domain generalization (DG), there is multiple training and test settings for each dataset, so the sample size indicates the entire dataset. ``DG" is short for domain generalization. ``\#" indicates the number or size. 
}
\label{tab:datainfo}
\resizebox{0.75\textwidth}{!}{%
\begin{tabular}{@{}c|c|c|ccc|c|c@{}}
\toprule
Source                    & Test Dataset   & \#Classes & \multicolumn{1}{c|}{\#Train Set}              & \multicolumn{1}{c|}{\#Val Set}              & \#Test Set & Shift Type      & \#Trained Models     \\ \midrule
\multirow{5}{*}{CIFAR-10} & CIFAR-10-C     & 10        & \multicolumn{1}{c|}{\multirow{5}{*}{50000}}   & \multicolumn{1}{c|}{\multirow{5}{*}{10000}} & 19*50000   & Corruption      & \multirow{5}{*}{57}  \\
                          & CIFAR-10.1     & 0         & \multicolumn{1}{c|}{}                         & \multicolumn{1}{c|}{}                       & 2021+2000  & Data collection &                      \\
                          & CIFAR-10.2     & 10        & \multicolumn{1}{c|}{}                         & \multicolumn{1}{c|}{}                       & 2000       & Data collection &                      \\
                          & CINIC-10       & 10        & \multicolumn{1}{c|}{}                         & \multicolumn{1}{c|}{}                       & 70000      & Data collection &                      \\
                          & STL-10         & 10    & \multicolumn{1}{c|}{}                         & \multicolumn{1}{c|}{}                       & 7200       & Data collection &                      \\ \midrule
CIFAR-100                 & CIFAR-100-C    & 100       & \multicolumn{1}{c|}{50000}                    & \multicolumn{1}{c|}{10000}                  & 19*50000   & Corruption      & 108                  \\ \midrule
\multirow{8}{*}{ImageNet} & TinyImageNet-C & 200       & \multicolumn{1}{c|}{\multirow{8}{*}{1281167}} & \multicolumn{1}{c|}{\multirow{8}{*}{50000}} & 15*5*10000 & Corruption      & \multirow{8}{*}{109} \\
                          & ImageNet-C     & 1000      & \multicolumn{1}{c|}{}                         & \multicolumn{1}{c|}{}                       & 19*5*50000 & Corruption      &                      \\
                          & ImageNet-S     & 1000      & \multicolumn{1}{c|}{}                         & \multicolumn{1}{c|}{}                       & 50889      & Style           &                      \\
                          & ImageNet-R     & 200       & \multicolumn{1}{c|}{}                         & \multicolumn{1}{c|}{}                       & 30000      & Style           &                      \\
                          & ObjectNet      & 313 & \multicolumn{1}{c|}{}                         & \multicolumn{1}{c|}{}                       & 18574      & Camera location &                      \\
                          & ImageNet-V2    & 1000      & \multicolumn{1}{c|}{}                         & \multicolumn{1}{c|}{}                       & 3*10000    & Data collection &                      \\
                          & ImageNet-A     & 200       & \multicolumn{1}{c|}{}                         & \multicolumn{1}{c|}{}                       & 7500       & Adversarial     &                      \\
                          & ImageNet-Vid   & 30        & \multicolumn{1}{c|}{}                         & \multicolumn{1}{c|}{}                       & 22179      & Time            &                      \\ \midrule
\multirow{2}{*}{WILDS}    & iWildCam       & 182       & \multicolumn{1}{c|}{129809}                   & \multicolumn{1}{c|}{7314}                   & 42791      & Camera location & 90                   \\
                          & FMoW           & 62        & \multicolumn{1}{c|}{76863}                    & \multicolumn{1}{c|}{11483}                  & 22108      & Time, region    & 90                   \\
                          & Camelyon17       & 2         & \multicolumn{1}{c|}{302436}                   & \multicolumn{1}{c|}{33560}                  & 85054      & Data collection & 90                   \\
                          & RxRx1          & 1139      & \multicolumn{1}{c|}{40612}                    & \multicolumn{1}{c|}{40612}                  & 34432      & Batch effect    & 90                   \\
                          & Amazon         & 5         & \multicolumn{1}{c|}{245502}                   & \multicolumn{1}{c|}{46950}                  & 100050     & Data collection & 39                   \\
                          & CivilComments  & 2         & \multicolumn{1}{c|}{269038}                   & \multicolumn{1}{c|}{45180}                  & 133782     & Demographics    & 30                   \\ \midrule
\multirow{6}{*}{DG}       & PACS           & 7         & \multicolumn{3}{c|}{9991}                                                                                & Style           & 120                  \\
                          & OfficeHome     & 65        & \multicolumn{3}{c|}{15588}                                                                               & Style           & 120                  \\
                          & DomainNet      & 345       & \multicolumn{3}{c|}{586920}                                                                              & Style           & 90                   \\
                          & NICO++         & 60        & \multicolumn{3}{c|}{88926}                                                                               & Background      & 90                   \\
                          & VLCS           & 5         & \multicolumn{3}{c|}{10729}                                                                               & Data collection & 120                  \\
                          & TerraInc       & 10        & \multicolumn{3}{c|}{24330}                                                                               & Camera location & 120                  \\ \midrule
\multirow{3}{*}{Subpop}   & Waterbirds     & 2         & \multicolumn{1}{c|}{4795}                     & \multicolumn{1}{c|}{1199}                   & 642        & Background      & 30                   \\
                          & CelebA         & 2         & \multicolumn{1}{c|}{162770}                   & \multicolumn{1}{c|}{19867}                  & 180        & Demographics    & 30                   \\
                          & CheXpert       & 2         & \multicolumn{1}{c|}{167093}                   & \multicolumn{1}{c|}{22280}                  & 661        & Demographics    & 30                   \\ \bottomrule
\end{tabular}%
}
\end{table*}

\section{Notations and Problem}
\label{sec:problem}

\paragraph{Notations} We use $X\in\mcX$ to denote input variables and $Y\in\mcY$ to denote the outcome, where $\mcX$ and $\mcY$ denote their support. 
$P^{tr}(X,Y)$ denotes training distribution and $P^{te}(X,Y)$ denotes test distribution. 
In the OOD circumstance, $P^{te}(X,Y)\neq P^{tr}(X,Y)$. 
A trained model is denoted as $f_{\theta_0}: \mcX\rightarrow\mcY$ with fixed parameters $\theta_0$. 
A validation dataset is denoted as $\{x_i^{va}, y_i^{va}\}_{i=1}^{n_{va}}$ and an OOD test dataset is denoted as $\{x_i^{te}, y_i^{te}\}_{i=1}^{n_{te}}$. 
Usually validation data follows the training distribution, i.e. $(x_i^{va}, y_i^{va})\sim P^{tr}(X,Y)$, and test data $(x_i^{te}, y_i^{te})\sim P^{te}(X,Y)$.

\paragraph{Problem} Given a trained model $f_{\theta_0}$, a labeled validation dataset $\{x_i^{va}, y_i^{va}\}_{i=1}^{n_{va}}$, and an unlabeled OOD test dataset $\{x_i^{te}\}_{i=1}^{n_{te}}$, the goal is to predict performance of the model on the test dataset so that it is close to ground truth performance, or to calculate a score positively correlated with ground truth performance. 
For a few algorithms of direct performance estimation, we can calculate the gap between estimated performances and ground truth performances for evaluation. 
For algorithms predicting a surrogate score instead of estimating the value of performance, since usually there are multiple trained models or multiple OOD test datasets, we can measure the correlation between surrogate scores and ground truth performances for evaluation.

%% file: sec/3_2_method.tex
\section{Benchmark Design}
\label{sec:benchmark}

In this section, we introduce the organization of our benchmark. More details are in~\Cref{appendix:detail}.

\begin{table*}[htbp]
\centering
\caption{Examples of representative datasets for various types of distribution shift. For the shift type of demographics, it represents demographic attributes like sex. For example, in training and validation data of CelebA, they mostly consist of blond hair female and black hair male, but in test data it mostly consists of blond hair male. }
\label{tab:example}
\resizebox{0.8\textwidth}{!}{%
\begin{tabular}{@{}c|c|ccl|lll@{}}
\toprule
Type            & Dataset    & \multicolumn{3}{c|}{Val} & \multicolumn{3}{c}{Test} \\ \midrule
Corruption      & ImageNet-C &        \raisebox{-.5\height}{\includegraphics[width=2cm]{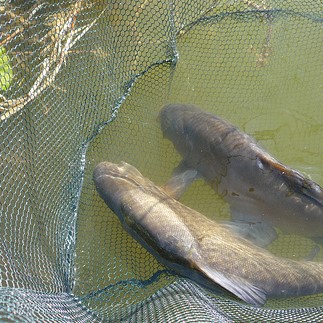}}     &  \raisebox{-.5\height}{\includegraphics[width=2cm]{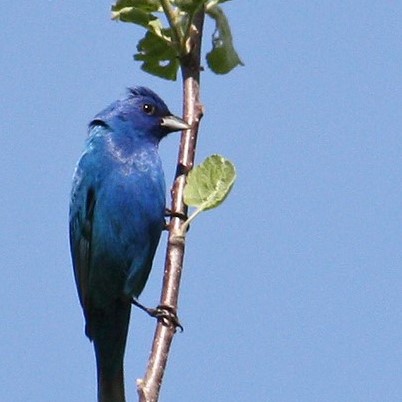}}      &  \raisebox{-.5\height}{\includegraphics[width=2cm]{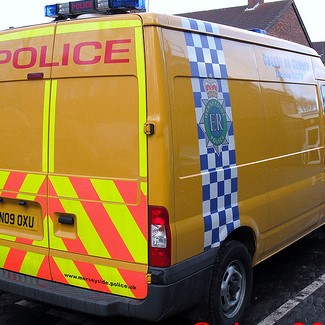}}      &  \raisebox{-.5\height}{\includegraphics[width=2cm]{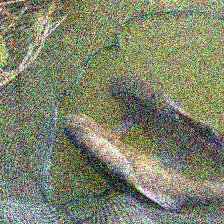}}      &  \raisebox{-.5\height}{\includegraphics[width=2cm]{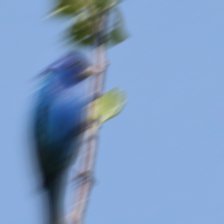}}      &  \raisebox{-.5\height}{\includegraphics[width=2cm]{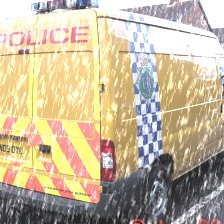}}         \\ \midrule
Style           & PACS       &   \raisebox{-.5\height}{\includegraphics[width=2cm]{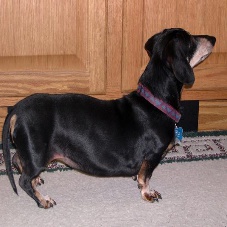}}     &  \raisebox{-.5\height}{\includegraphics[width=2cm]{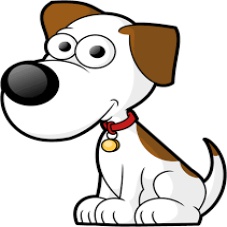}}      &  \raisebox{-.5\height}{\includegraphics[width=2cm]{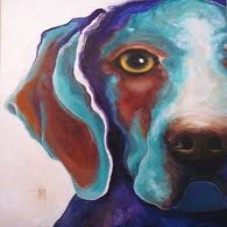}}      &  \raisebox{-.5\height}{\includegraphics[width=2cm]{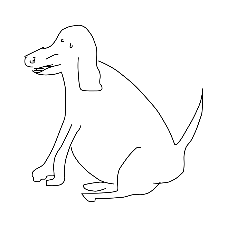}}      &  \raisebox{-.5\height}{\includegraphics[width=2cm]{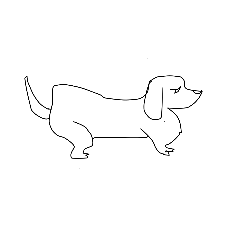}}      &  \raisebox{-.5\height}{\includegraphics[width=2cm]{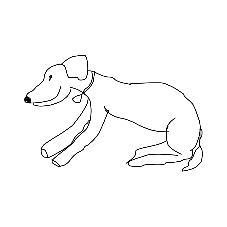}}    \\ \midrule
Background      & NICO++     &   \raisebox{-.5\height}{\includegraphics[width=2cm]{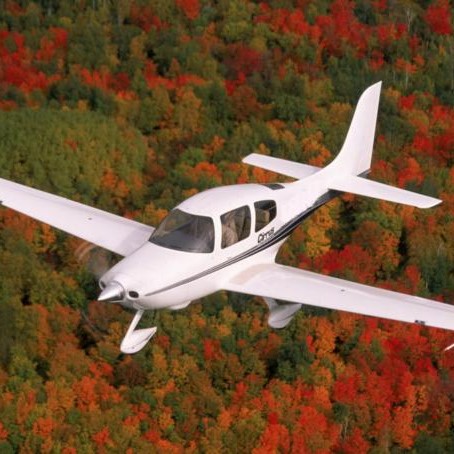}}     &  \raisebox{-.5\height}{\includegraphics[width=2cm]{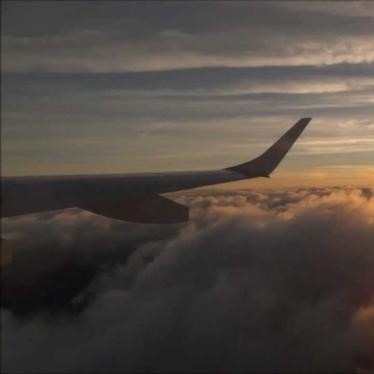}}      &  \raisebox{-.5\height}{\includegraphics[width=2cm]{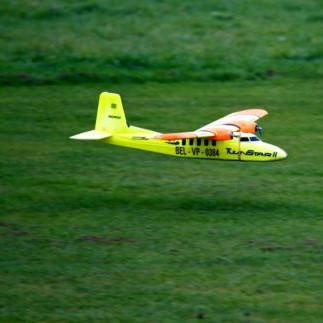}}      &  \raisebox{-.5\height}{\includegraphics[width=2cm]{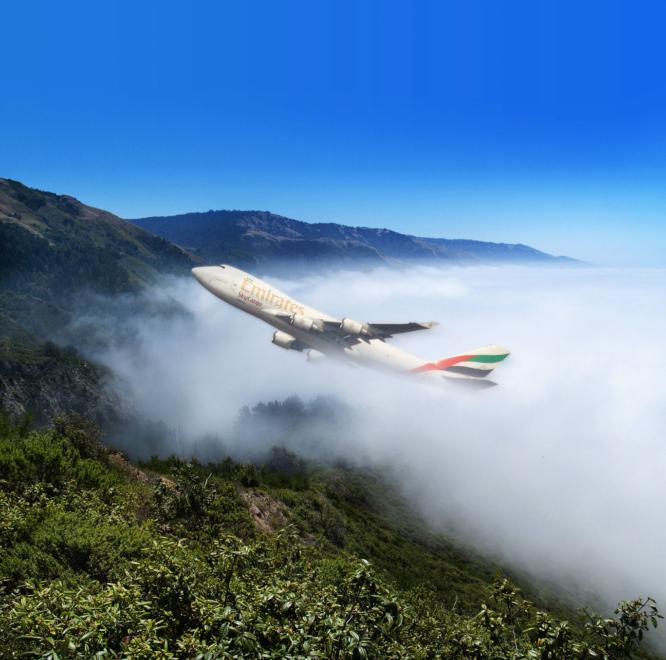}}      &  \raisebox{-.5\height}{\includegraphics[width=2cm]{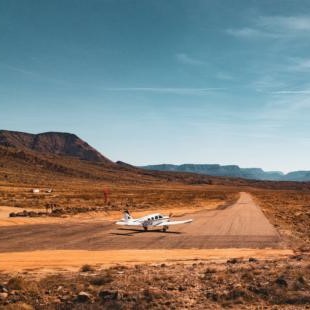}}      &  \raisebox{-.5\height}{\includegraphics[width=2cm]{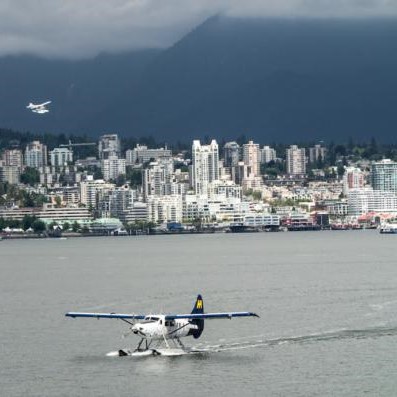}}    \\ \midrule
Data collection & VLCS       &   \raisebox{-.5\height}{\includegraphics[width=2cm]{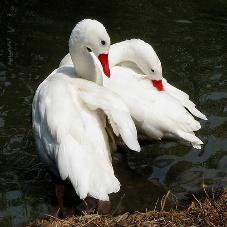}}     &  \raisebox{-.5\height}{\includegraphics[width=2cm]{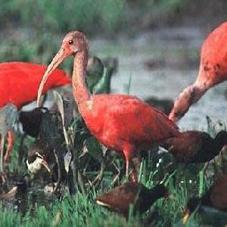}}      &  \raisebox{-.5\height}{\includegraphics[width=2cm]{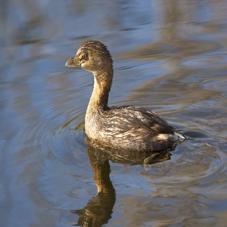}}      &  \raisebox{-.5\height}{\includegraphics[width=2cm]{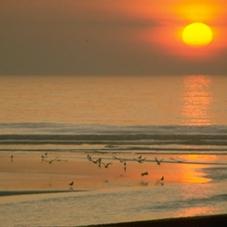}}      &  \raisebox{-.5\height}{\includegraphics[width=2cm]{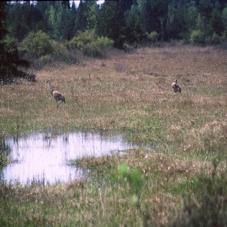}}      &  \raisebox{-.5\height}{\includegraphics[width=2cm]{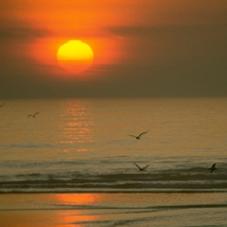}}      \\ \midrule
Camera location & iWildCam   & \raisebox{-.5\height}{\includegraphics[width=2cm]{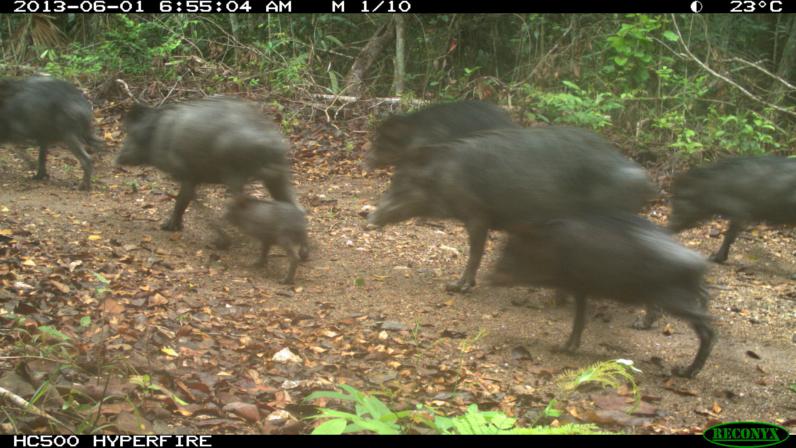}}     &  \raisebox{-.5\height}{\includegraphics[width=2cm]{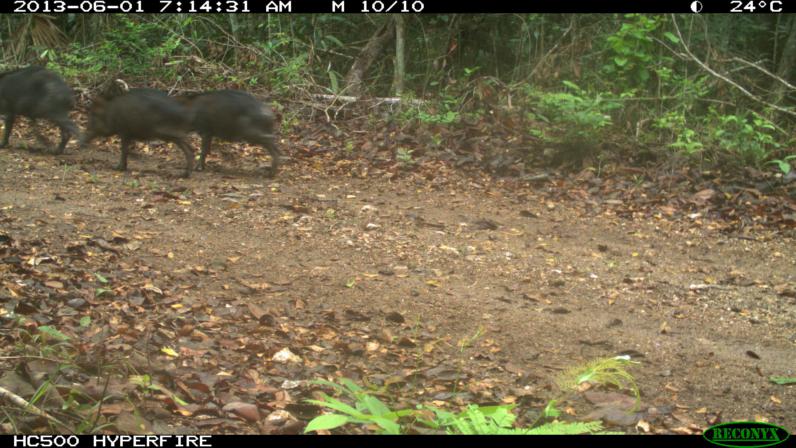}}      &  \raisebox{-.5\height}{\includegraphics[width=2cm]{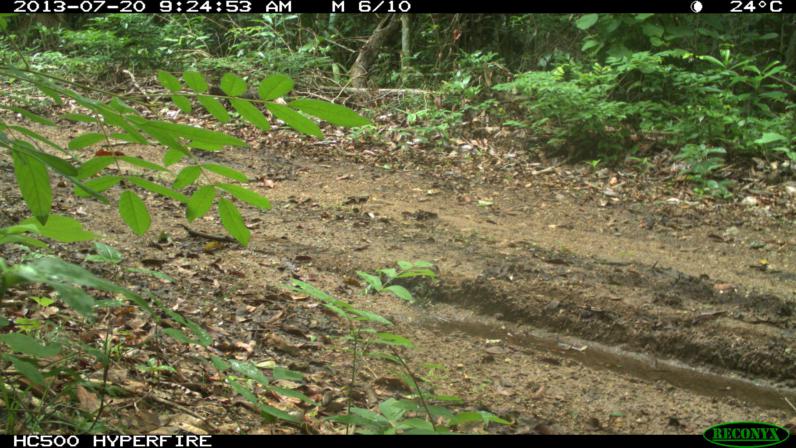}}      &  \raisebox{-.5\height}{\includegraphics[width=2cm]{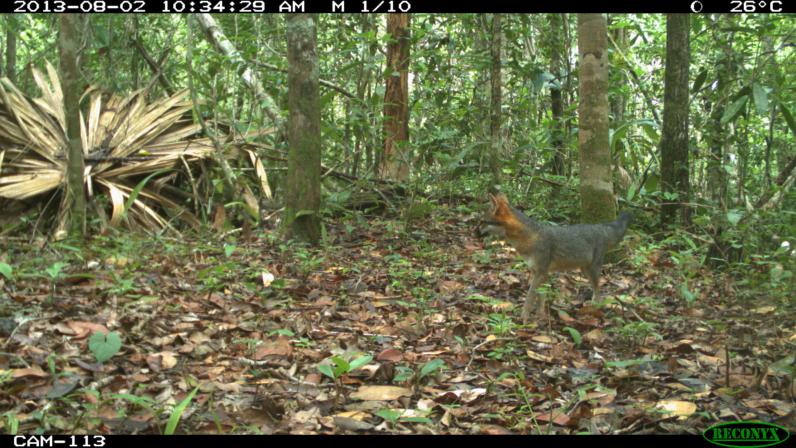}}      &  \raisebox{-.5\height}{\includegraphics[width=2cm]{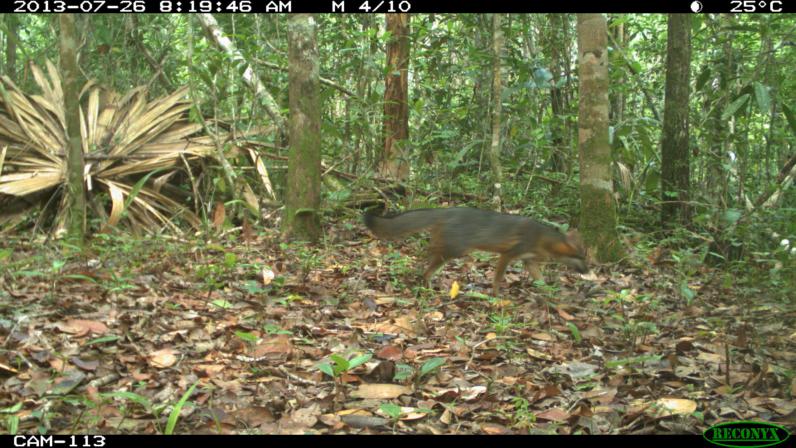}}      &  \raisebox{-.5\height}{\includegraphics[width=2cm]{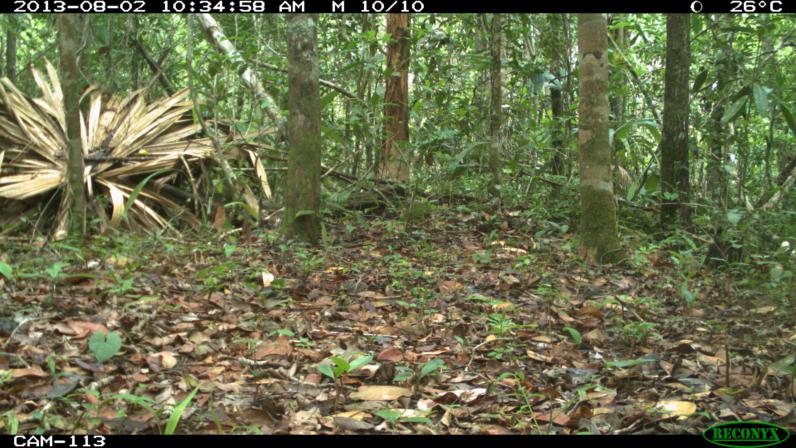}}    \\ \midrule
Demographics    & CelebA     &   \raisebox{-.5\height}{\includegraphics[width=2cm]{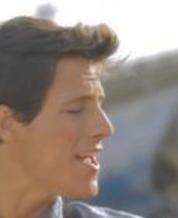}}     &  \raisebox{-.5\height}{\includegraphics[width=2cm]{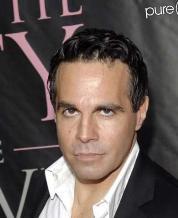}}      &  \raisebox{-.5\height}{\includegraphics[width=2cm]{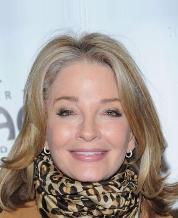}}      &  \raisebox{-.5\height}{\includegraphics[width=2cm]{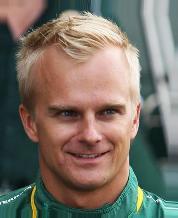}}      &  \raisebox{-.5\height}{\includegraphics[width=2cm]{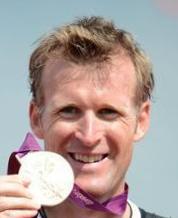}}      &  \raisebox{-.5\height}{\includegraphics[width=2cm]{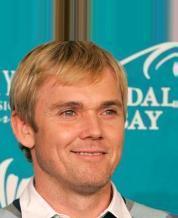}}      \\ \bottomrule
\end{tabular}%
}
\end{table*}

\subsection{Datasets}

In previous works of OOD performance prediction, they only evaluate on a limited number of datasets, and they mostly focus on the case where ImageNet and CIFAR-10 are treated as the training dataset. 
They seldom evaluate performance prediction algorithms on OOD datasets that are widely adopted in domain generalization and subpopulation shift. 
Besides, the types of distribution shifts previous works have covered are not diverse enough.
They have mostly covered shifts of synthetic corruptions, style, and the process of data collection, but they seldom investigate shifts caused by camera locations, image backgrounds, and demographic attributes. 
To establish a comprehensive benchmark, we include \NUMDATA~OOD datasets in our benchmark, covering most commonly used datasets in the area of OOD generalization and OOD performance prediction and more diverse types of distribution shifts. 
We list detailed information of the included OOD datasets in~\Cref{tab:datainfo}. 
We also provide examples from representative OOD datasets for common types of shifts in~\Cref{tab:example}.
For variants of CIFAR~\citep{krizhevsky2009learning}, we include CIFAR-10-C and CIFAR-100-C~\citep{hendrycks2019benchmarking} as synthetic distribution shifts induced by corruptions, and we include CIFAR-10.1~\citep{recht2018cifar}, CIFAR-10.2~\citep{lu2020harder}, CINIC-10~\citep{darlow2018cinic}, and STL-10~\citep{coates2011analysis} as representatives of real-world shifts. 
For variants of ImageNet~\citep{deng2009imagenet}, we include ImageNet-C and TinyImageNet-C~\citep{hendrycks2019benchmarking}  as synthetic distribution shifts also caused by corruptions, and we include ImageNet-V2~\citep{recht2019imagenet}, ImageNet-S~\citep{wang2019learning}, ImageNet-A~\citep{hendrycks2021natural}, ImageNet-R~\citep{hendrycks2021many}, ObjectNet~\citep{barbu2019objectnet}, and ImageNet-Vid-Robust~\citep{shankar2019systematic} covering various types of real-world shifts. 
For WILDS~\citep{koh2021wilds}, we choose iWildCam~\citep{beery2021iwildcam}, FMoW~\citep{christie2018functional}, Camelyon17~\citep{bandi2018detection}, RxRx1~\citep{sypetkowski2023rxrx1}, Amazon~\citep{ni2019justifying}, and CivilComments~\citep{borkan2019nuanced} as representative datasets. 
For domain generalization, we include all commonly used ones: PACS~\citep{li2017deeper}, VLCS~\citep{khosla2012undoing}, OfficeHome~\citep{venkateswara2017deep}, TerraInc~\citep{beery2018recognition}, DomainNet~\citep{peng2019moment}, and NICO++~\citep{zhang2023nico++}. 
For subpopulation shift, we include three most commonly used ones: Waterbirds~\citep{sagawa2020distributionally}, CelebA~\citep{liu2015deep}, and CheXpert~\citep{irvin2019chexpert}. 
Note that for variants of ImageNet and CIFAR, and datasets from WILDS, each dataset corresponds to only one fixed setting of training and testing, while for domain generalization datasets, each employs leave-one-domain/group-out practice, i.e. testing on one domain or one group of domains and training on the rest domains, thus leading to multiple settings. 
For subpopulation shift, we treat the worst subgroup as the OOD test dataset, which is fixed and determined in advance according to experiments in previous literature.

\subsection{Training}
\label{sec:training}

We provide a total of \NUMMODEL~off-the-shelf trained models for subsequent researchers to directly test their performance prediction algorithms on. Note that this not only enables fair comparisons between algorithms on the same testbench, but also greatly reduces their burden of training models repeatedly. 
The number of trained models for each dataset can also be referred to in~\Cref{tab:datainfo}. 
For variants of ImageNet, we directly leverage 109 open-sourced models of different architectures included in Torchvision.
For variants of CIFAR-10 and CIFAR-100, we train from scratch with each architecture three random seeds following default settings in two public repositories\footnote{https://github.com/kuangliu/pytorch-cifar}\footnote{https://github.com/weiaicunzai/pytorch-cifar100}. 
The former yields $19\times 3$ models while the latter $36\times 3$ models. 
For WILDS, we follow its default setting~\citep{koh2021wilds} to train one model initialized from ImageNet supervised pretrained weights for 30 different model architectures. 
For domain generalization datasets, we follow~\citet{gulrajani2020search} and~\citet{yu2024rethinking} to adopt the leave-one-domain-out setting for PACS, VLCS, OfficeHome, and TerraInc, and the leave-one-group-out setting for DomainNet and NICO++, and we use supervised, MoCo-v2~\citep{chen2020improved}, and CLIP~\citep{radford2021learning} pretrained weights as initialization for ResNet-50~\citep{he2016deep}, and supervised, MoCo-v3~\citep{chen2021empirical}, and CLIP pretrained weights as initialization for ViT-B/16~\citep{dosovitskiy2021image}. Each setting leads to 5 models with different random seeds, yielding 30 trained models. 
For subpopulation shift datasets, we employ the same training protocol as that of domain generalization, thus each also yielding 30 trained models. 

\subsection{Algorithms}
In our benchmark, we evaluate \NUMALG~performance prediction algorithms, covering almost all practical ones. We do not include algorithms that require additional training of models~\citep{deng2021does, chen2021detecting, yu2022predicting}. 
We list adopted algorithms below:
\begin{itemize}
    \item Average Thresholded Confidence (ATC)~\citep{garg2022leveraging}: Learn a threshold on model confidence and employ fraction of samples with higher confidence as the predicted accuracy. 
    \item Difference of Confidences (DoC)~\citep{guillory2021predicting}: Calculate the difference between average model confidence on the OOD test dataset and the validation dataset.
    \item Nuclear Norm~\citep{deng2023confidence}: Adopt nuclear norm of prediction matrix to measure both confidence and dispersity. 
    \item Neighborhood Invariance (NI)~\citep{ng2024predicting}: Calculate the label invariance across augmented examples in the neighborhood of a given test sample. 
    \item Matrix Norm (MaNo)~\citep{xie2024mano}: Calculate the $L_p$ norm of the prediction matrix. 
    \item Dispersion Score (Dispersion)~\citep{xie2023importance}: Measure inter-class feature separability with average distances between each feature cluster centroid induced by pseudo labeling and the centroid of all features. 
    \item Meta-Distribution Energy (MDE)~\citep{peng2024energy}: Calculate average energy of unlabeled test data via probability density of the Gibbs distribution. 
    \item Agreement~\citep{baek2022agreement}: Leverage agreement-on-the-line to predict model performance based on model agreement. 
    \item Confidence Optimal Transport (COT)~\citep{lu2023characterizing}: Calculate Wasserstein distance between predicted label distribution on test data and true label distribution on validation data. 
    \item Confidence Optimal Transport with Thresholding (COTT)~\citep{lu2023characterizing}: A variant of COT that applies thresholding to transport costs to improve estimation. 
\end{itemize}

\subsection{Evaluation Metrics}

In previous literature, only a few methods are capable of directly estimating the value of accuracy. Actually, there is usually a series of trained models or OOD datasets available. Thus it is generally practical enough to provide a series of surrogate scores so that the predicted best-performing model or dataset can be selected. 
Therefore, for each OOD dataset, we calculate \textbf{Spearman's rank correlation $\rho$} between predicted scores and ground truth performances of a set of trained models to evaluate the effectiveness of performance prediction algorithms: 
\begin{equation}
    \rho = 1 - \frac{6 \sum_{i=1}^n (R(\hat{S}_i)-R(Acc_i))^2}{n(n^2 - 1)}
\end{equation}
Where $\hat{S}_i$ represents the $i$-th predicted score, $Acc_i$ represents the $i$-th ground truth accuracy, and $R(\cdot)$ implies the rank of $\hat{S}_i$ or $Acc_i$ in their sequence. 
This metric has been widely adopted before~\citep{baek2022agreement, guillory2021predicting,xie2023importance,deng2023confidence,xie2024mano, xie2024gradient, peng2024energy}. 
We do not choose coefficient of determination $R^2$ as the main evaluation metric since it is quite sensitive to outliers and it cannot address the nonlinear correlation. 
We will conduct analyses of $R^2$ and $\rho$ in \Cref{sec:metric}. 
Besides, some previous works calculate $R^2$ and $\rho$ of models only when the models share the same architecture, and most of the results are larger than 0.95, even quite close to 1~\citep{xie2023importance,deng2023confidence,xie2024mano, xie2024gradient, peng2024energy}. 
Although this demonstrates the effectiveness of the proposed algorithms to some extent, it might not be practical enough since there are usually different architectures in a given pool of trained models. 
Thus in our benchmark \OURS, we calculate these metrics for models across different architectures, which is both more challenging and more meaningful.

%% file: sec/4_exp.tex
\section{Experiments}
\label{sec:exp}

In this section, we present detailed results of our established benchmark \OURS~and conduct other experimental analyses of evaluation metrics, model architectures, pretrained weights, etc. 

\begin{table*}[htbp]
\centering
\caption{Complete results of \OURS including \NUMDATA~OOD datasets and \NUMALG~OOD performance prediction algorithms. The adopted metric is Spearman’s rank correlation between predicted scores and ground truth performances. The last column represents the number of effective algorithms on the corresponding dataset, where ``effective" is defined as $\rho>0.7$. 
}
\label{tab:benchmark}
\resizebox{\textwidth}{!}{%
\begin{tabular}{@{}c|cccccccccc|cc@{}}
\toprule
Dataset        & ATC    & Nu. Norm & DOC    & NI     & MaNo   & Dispersion & MDE    & Agreement & COT    & COTT   & Avg.   & \#Effective \\ \midrule
CIFAR-10-C     & 0.951  & 0.934    & 0.884  & 0.818  & 0.836  & 0.903      & -0.832 & 0.991     & 0.980  & 0.990  & 0.746  & 9           \\ \midrule
CIFAR-100-C    & 0.948  & 0.829    & 0.874  & 0.868  & 0.738  & 0.718      & -0.703 & 0.971     & 0.892  & 0.985  & 0.712  & 9           \\ \midrule
NICO++         & 0.970  & 0.797    & 0.742  & 0.929  & 0.846  & 0.802      & -0.777 & 0.932     & 0.844  & 0.965  & 0.705  & 9           \\ \midrule
OfficeHome     & 0.911  & 0.594    & 0.465  & 0.912  & 0.754  & 0.541      & -0.409 & 0.936     & 0.693  & 0.933  & 0.633  & 5           \\ \midrule
ObjectNet      & 0.794  & 0.674    & 0.640  & 0.875  & 0.672  & 0.689      & -0.670 & 0.973     & 0.706  & 0.845  & 0.620  & 5           \\ \midrule
ImageNet-C     & 0.933  & 0.596    & 0.556  & 0.790  & 0.453  & 0.581      & -0.506 & 0.933     & 0.583  & 0.839  & 0.576  & 4           \\ \midrule
ImageNet-V2    & 0.993  & 0.260    & 0.958  & 0.845  & 0.149  & 0.323      & -0.216 & 0.996     & 0.357  & 0.986  & 0.565  & 5           \\ \midrule
ImageNet-S     & 0.961  & 0.500    & 0.480  & 0.846  & 0.343  & 0.438      & -0.384 & 0.838     & 0.582  & 0.987  & 0.559  & 4           \\ \midrule
PACS           & 0.708  & 0.769    & 0.657  & 0.825  & 0.489  & 0.772      & -0.736 & 0.909     & 0.498  & 0.636  & 0.553  & 5           \\ \midrule
CIFAR-10.1     & 0.717  & 0.526    & 0.423  & 0.766  & 0.524  & 0.533      & -0.524 & 0.850     & 0.556  & 0.718  & 0.509  & 4           \\ \midrule
CIFAR-10.2     & 0.780  & 0.511    & 0.364  & 0.800  & 0.511  & 0.511      & -0.508 & 0.807     & 0.541  & 0.750  & 0.507  & 4           \\ \midrule
ImageNet-R     & 0.742  & 0.325    & 0.338  & 0.859  & 0.414  & 0.232      & -0.278 & 0.926     & 0.404  & 0.885  & 0.485  & 4           \\ \midrule
TinyImageNet-C & 0.617  & 0.553    & 0.454  & 0.530  & 0.487  & 0.274      & -0.343 & 0.740     & 0.583  & 0.839  & 0.473  & 2           \\ \midrule
DomainNet      & 0.841  & 0.563    & 0.344  & 0.235  & 0.624  & 0.563      & -0.565 & 0.433     & 0.601  & 0.874  & 0.451  & 2           \\ \midrule
STL-10         & 0.718  & 0.432    & 0.003  & 0.825  & 0.412  & 0.419      & -0.402 & 0.771     & 0.567  & 0.745  & 0.449  & 4           \\ \midrule
FMoW           & 0.964  & 0.716    & 0.172  & 0.299  & -0.305 & 0.678      & -0.674 & 0.859     & 0.681  & 0.955  & 0.435  & 4           \\ \midrule
CINIC-10       & 0.706  & 0.318    & -0.075 & 0.867  & 0.320  & 0.307      & -0.301 & 0.794     & 0.423  & 0.728  & 0.409  & 4           \\ \midrule
VLCS           & 0.472  & -0.036   & 0.142  & 0.699  & 0.520  & 0.145      & 0.060  & 0.777     & 0.112  & 0.457  & 0.335  & 1           \\ \midrule
iWildCam       & 0.669  & 0.103    & 0.341  & 0.540  & 0.023  & 0.273      & 0.012  & 0.508     & 0.190  & 0.618  & 0.328  & 0           \\ \midrule
ImageNet-Vid   & -0.095 & 0.712    & -0.690 & 0.556  & 0.606  & 0.648      & -0.597 & 0.654     & 0.743  & 0.737  & 0.328  & 3           \\ \midrule
RxRx1          & 0.983  & 0.882    & -0.738 & -0.538 & -0.356 & 0.863      & -0.888 & 0.979     & 0.885  & 0.979  & 0.305  & 6           \\ \midrule
ImageNet-A     & 0.503  & 0.135    & 0.208  & 0.818  & 0.209  & 0.143      & -0.158 & 0.197     & 0.122  & 0.530  & 0.270  & 1           \\ \midrule
Amazon         & 0.749  & 0.077    & -0.225 & -      & 0.091  & 0.073      & -0.131 & 0.810      & 0.123  & 0.640   & 0.245  & 3           \\ \midrule
Camelyon17     & 0.296  & 0.632    & -0.506 & -0.513 & 0.056  & 0.747      & 0.315  & -0.637    & 0.796  & 0.668  & 0.185  & 2           \\ \midrule
TerraInc       & 0.370  & 0.440    & 0.282  & 0.355  & -0.067 & 0.438      & -0.482 & 0.422     & -0.122 & -0.062 & 0.157  & 0           \\ \midrule
CivilComments  & 0.806  & -0.779   & -0.096 & -      & 0.454  & -0.850      & 0.496  & 0.818     & -0.350  & 0.537  & 0.115  & 3           \\ \midrule
Waterbirds     & 0.219  & 0.449    & 0.825  & -0.194 & 0.149  & 0.475      & 0.037  & 0.473     & -0.960 & -0.883 & 0.059  & 1           \\ \midrule
CelebA         & 0.392  & 0.498    & 0.369  & 0.027  & 0.463  & 0.602      & -0.350 & 0.162     & -0.922 & -0.825 & 0.042  & 0           \\ \midrule
CheXpert       & 0.052  & 0.327    & -0.752 & -0.786 & -0.472 & 0.782      & 0.535  & -0.530    & -0.786 & -0.853 & -0.248 & 1           \\ \bottomrule
\end{tabular}%
}
\end{table*}

\subsection{Benchmark results}
\label{sec:mainexp}

The complete results are shown in \Cref{tab:benchmark}, where the datasets are sorted in a descending order of average rank correlations achieved by the \NUMALG~algorithms. 
The last column is the number of effective algorithms for a specific OOD dataset, where we defined ``effective`` as $\rho>0.7$. 
We can see that many algorithms achieve a considerable rank correlation on OOD datasets whose shifts are created by synthetic corruptions, including CIFAR-10-C, CIFAR-100-C, and ImageNet-C. These are exactly the most commonly adopted datasets in previous literature. 
This indicates that it would be better to shift focus to more complex real-world distribution shifts when evaluating performance prediction algorithms in the future. 
For OOD datasets whose shifts are induced by image style, effectiveness of current algorithms gradually decrease with the increasing complexity and diversity of styles. 
For PACS and OfficeHome that have four different styles, about half of algorithms achieve satisfying results. 
For DomainNet with six styles, only two algorithms exhibit effectiveness. 
For OOD datasets whose shifts come from the process of data collection, including CINIC-10, STL-10, CIFAR-10.1/10.2, and VLCS, fewer than half of algorithms work well. This might be due to the complexity of data collection process. 
Nevertheless, for NICO++ composed of six different domains whose shifts are generated by image background, the shifts should have been complex enough, while almost all algorithms are capable of achieving a high rank correlation. This could be because the complex shifts of background contribute mostly to the covariate shift, i.e. shift caused by $P(X)$ instead of concept shift, i.e. shift caused by $P(Y|X)$ in NICO++, since \citet{zhang2023nico++} find that NICO++ has smaller concept shift and larger covariate shift compared with other domain generalization datasets. 
In addition, it is worth noting that all three subpopulation datasets seem to be extremely ``hard" with no or only one algorithm achieving a rank correlation higher than 0.7. Many algorithms even achieve a negative rank correlation on these datasets. 
This might be due to over-confidence in certain subpopulations. For example, in CelebA, ERM-trained models tend to classify faces of blond hair male as not blond hair with high confidence since it learns the strong spurious correlation between blond hair and female. 
This indicates that more attention could be paid to performance prediction associated with subpopulation shifts.

\begin{figure*}[t]
	\centering
	\begin{subfigure}{0.33\linewidth}  
	    \includegraphics[width=\linewidth]{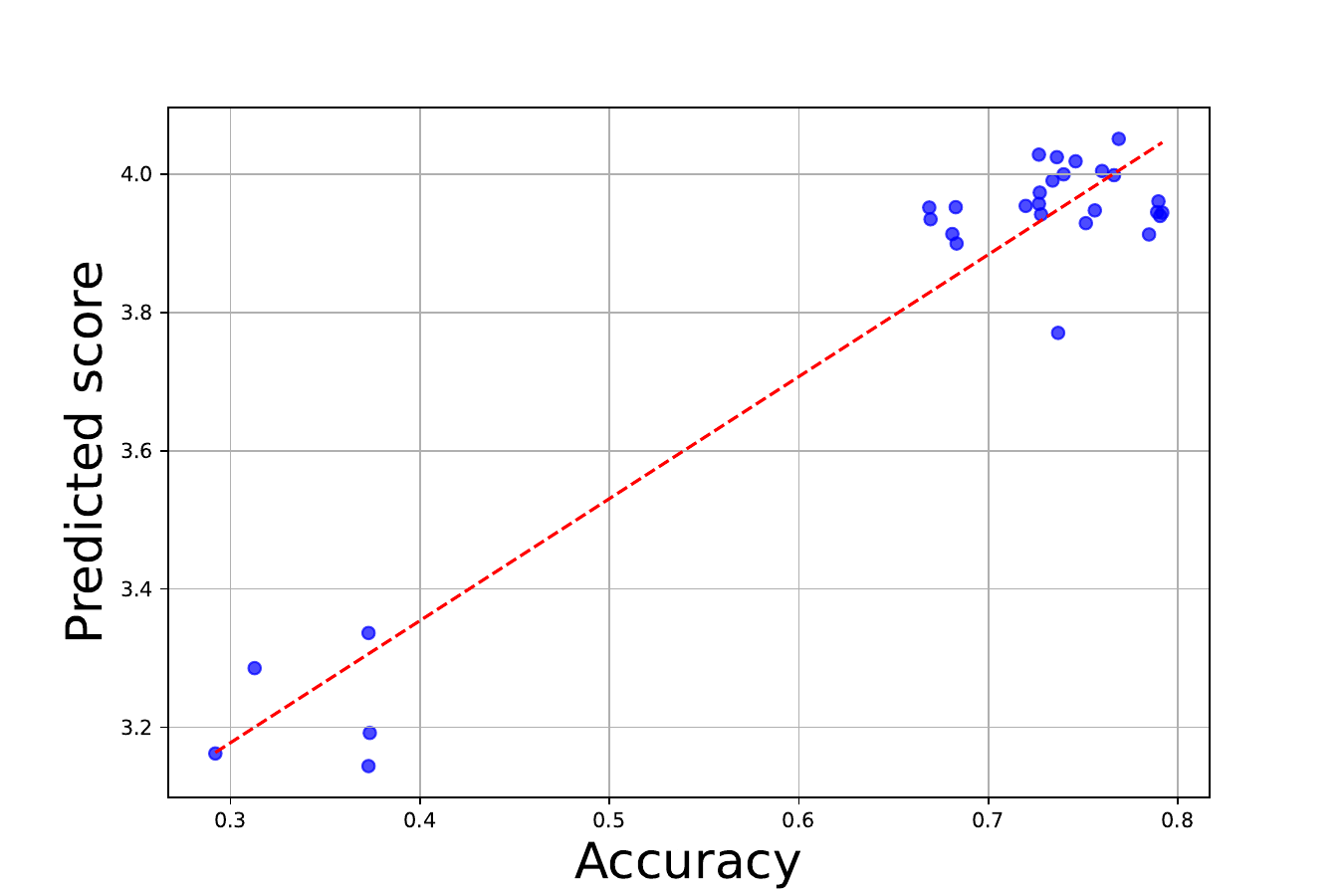}
        \caption{MaNo applied on Real domain of OfficeHome, where $R^2=0.912,\ \rho=0.475$.}
	\label{fig:pattern1}
	\end{subfigure}
	\begin{subfigure}{0.33\linewidth}  
	    \includegraphics[width=\linewidth]{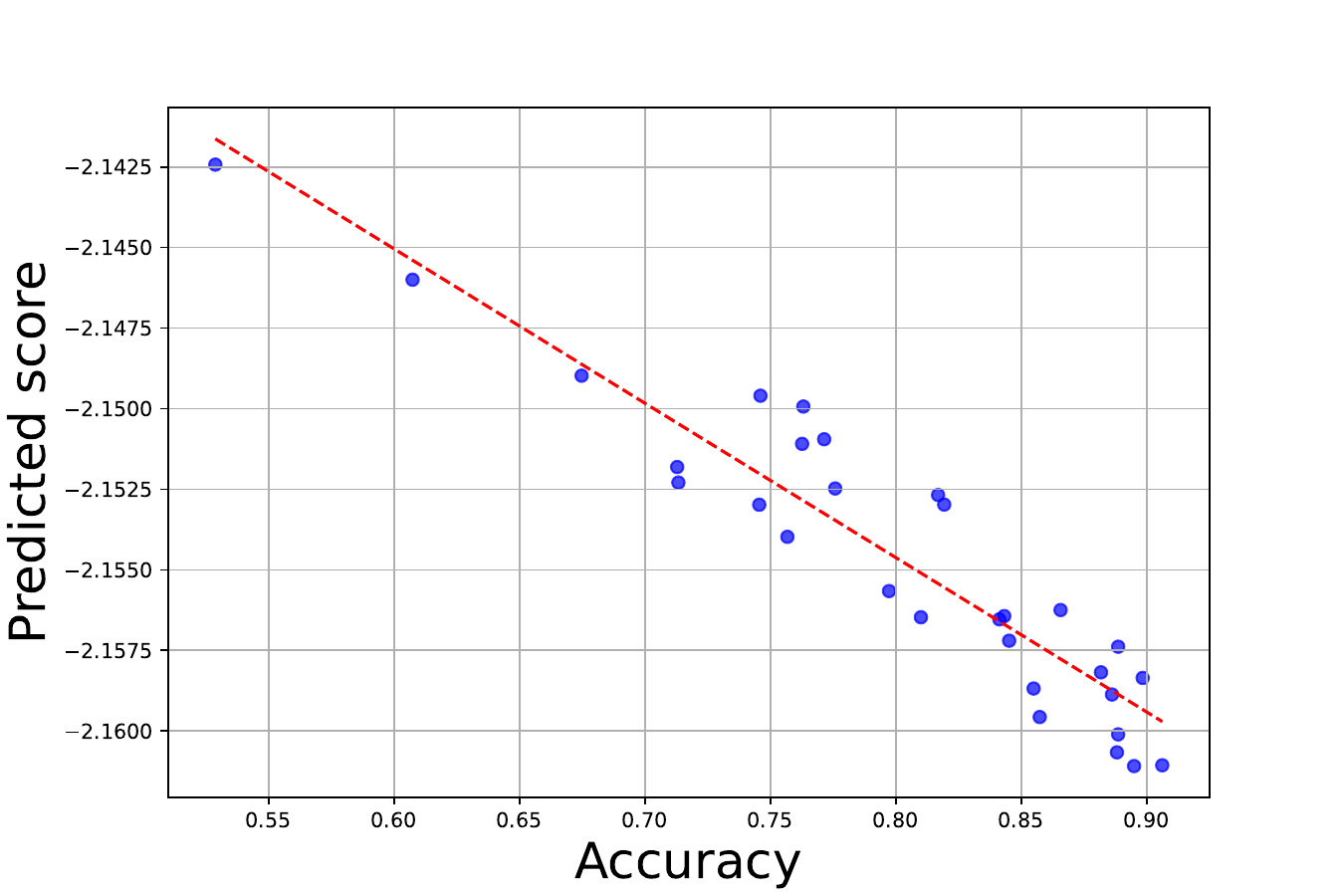}
        \caption{MDE applied on Art domain of PACS, where $R^2=0.881,\ \rho=-0.911$.}
	\label{fig:pattern2}
	\end{subfigure}
	\begin{subfigure}{0.33\linewidth}  
	    \includegraphics[width=\linewidth]{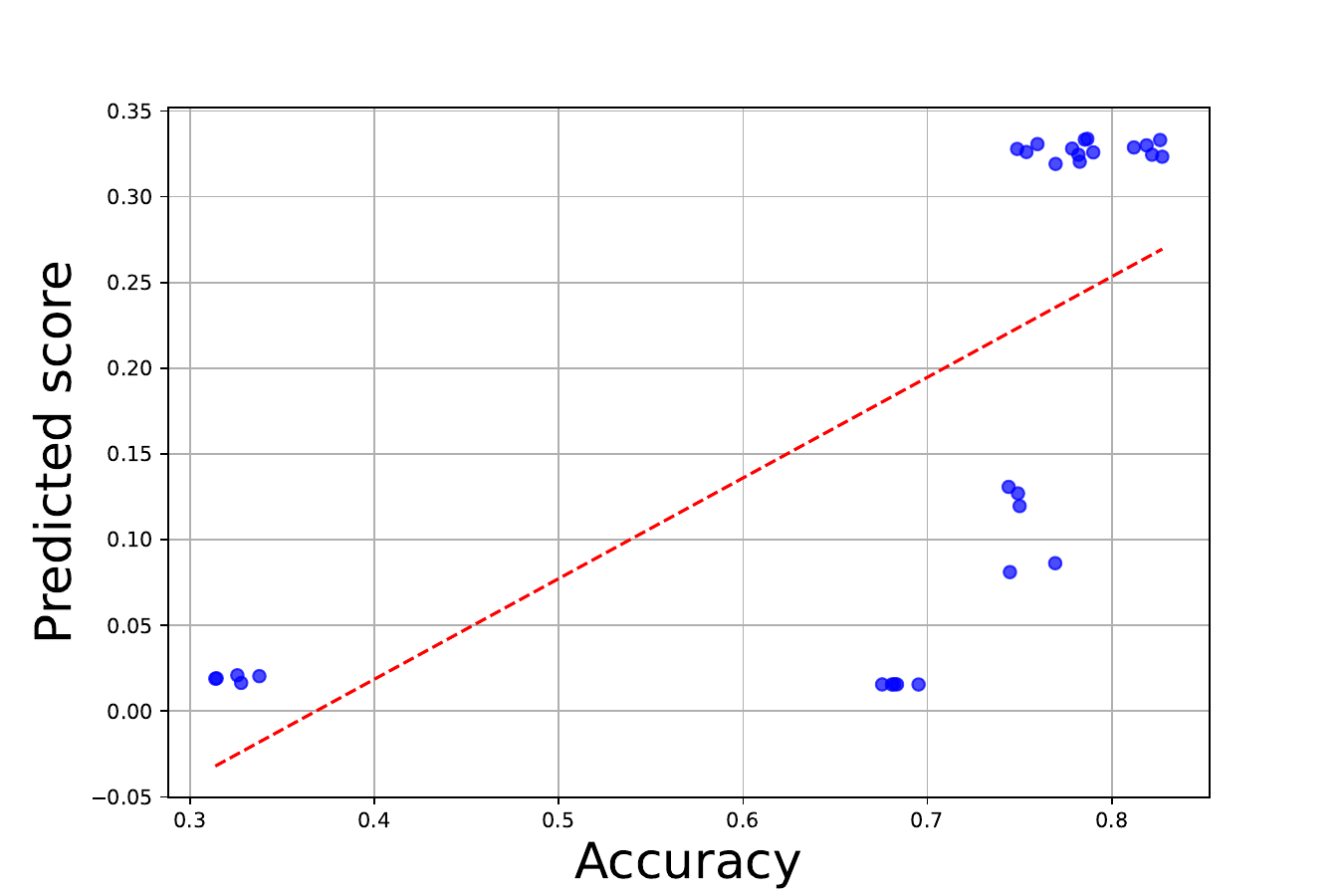}
        \caption{Dispersion applied on Product domain of OfficeHome, where $R^2=0.475,\ \rho=0.793$.}
	\label{fig:pattern3}
	\end{subfigure}
	\caption{Scatter plots of predicted scores against ground truth accuracies. 
    }
	\label{fig:metric}
\end{figure*}

\begin{table*}[htbp]
\centering
\caption{Rank correlation measured on PACS with models trained by different OOD generalization algorithms. Overall, performance prediction algorithms achieve slightly higher rank correlation for models trained with OOD generalization algorithms than with ERM.}
\label{tab:algorithm}
\resizebox{0.9\textwidth}{!}{%
\begin{tabular}{@{}c|cccccccccc|c@{}}
\toprule
OOD   Algorithm & ATC   & Nu. Norm & DOC   & NI    & MaNo  & Dispersion & MDE    & Agreement & COT   & COTT  & Avg.\\ \midrule
ERM             & 0.708 & 0.769       & 0.657 & 0.825 & 0.489 & 0.772      & -0.736 & 0.909     & 0.498 & 0.636 & 0.553   \\
RSC             & 0.744 & 0.762       & 0.653 & 0.728 & 0.788 & 0.778      & -0.716 & 0.893     & 0.534 & 0.785 & 0.595   \\
Mixup           & 0.784 & 0.758       & 0.678 & 0.798 & 0.610 & 0.766      & -0.687 & 0.854     & 0.804 & 0.831 & 0.620   \\
SWAD            & 0.698 & 0.730       & 0.642 & 0.778 & 0.589 & 0.715      & -0.631 & 0.856     & 0.605 & 0.704 & 0.569   \\
CORAL           & 0.740 & 0.719       & 0.672 & 0.740 & 0.460 & 0.729      & -0.619 & 0.885     & 0.599 & 0.770 & 0.569   \\ \bottomrule
\end{tabular}%
}
\end{table*}

\begin{figure*}[t]
	\centering
	\begin{subfigure}{0.32\linewidth}  
	    \includegraphics[width=\linewidth]{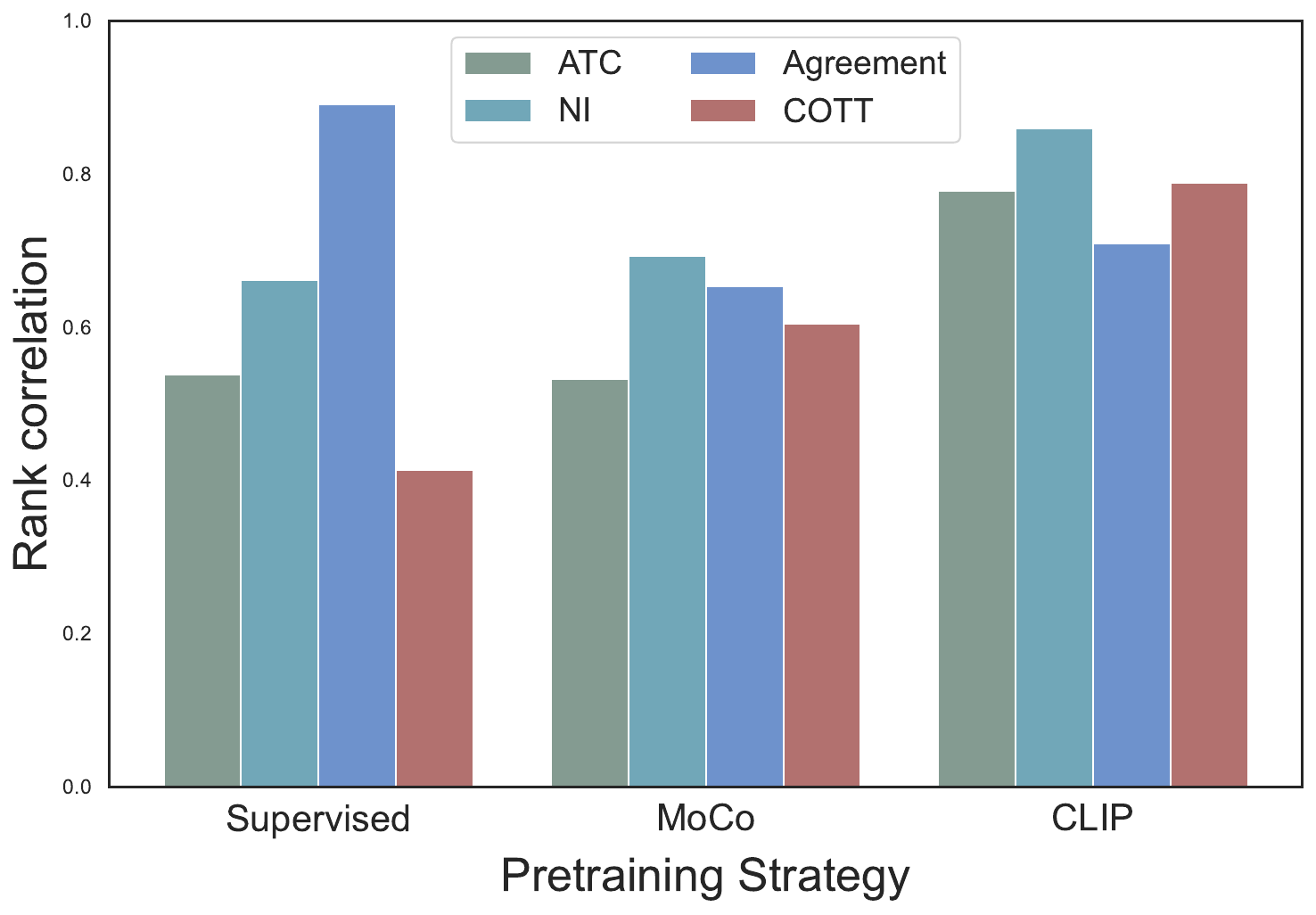}
        \caption{Rank correlation measured on PACS with models of varying pretraining strategies. 
        }
	\label{fig:pretrain}
	\end{subfigure}
	\begin{subfigure}{0.28\linewidth}  
	     \includegraphics[width=\linewidth]{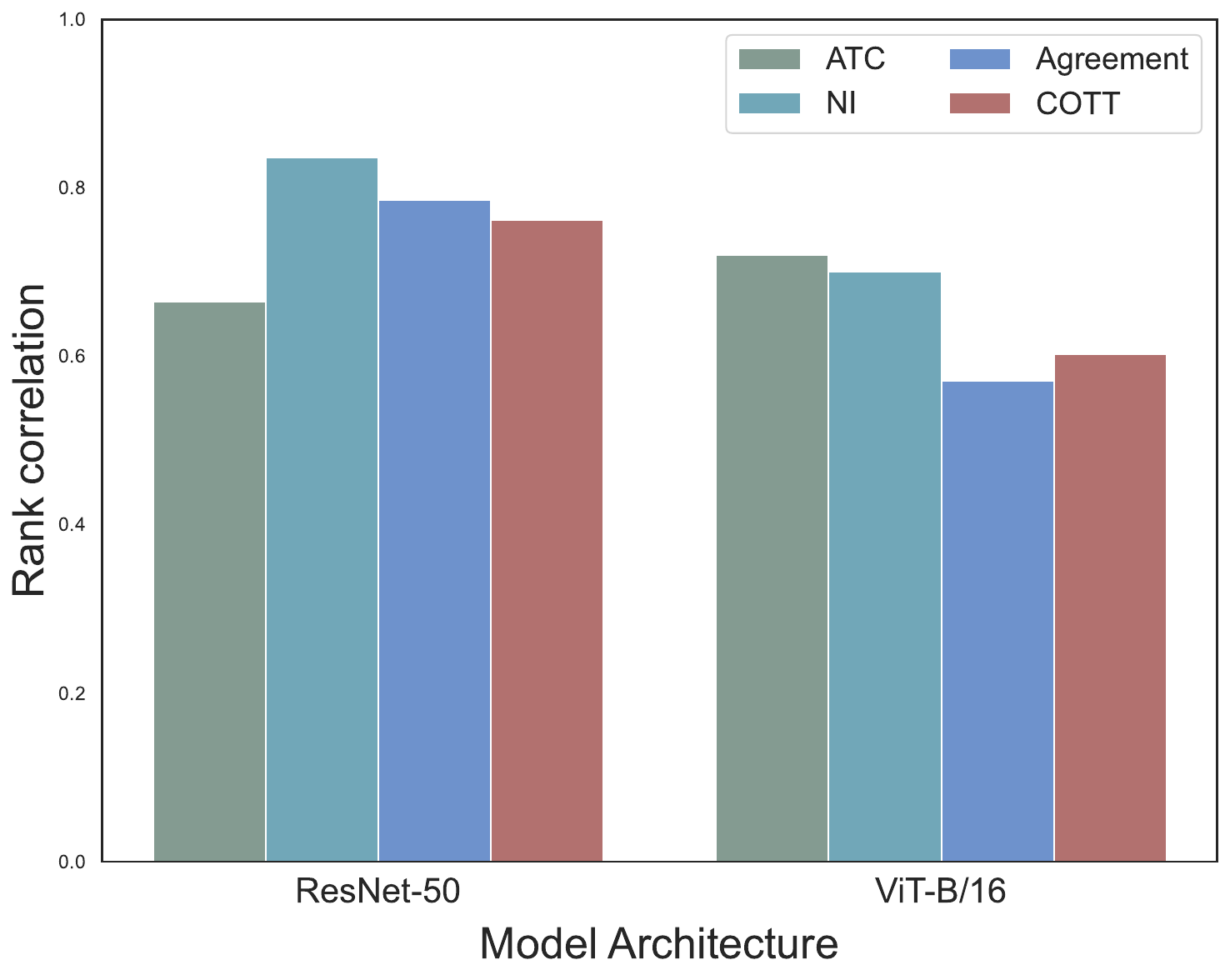}
        \caption{Rank correlation measured on PACS with models of varying architectures. 
        }
	\label{fig:arch}
	\end{subfigure}
	\begin{subfigure}{0.33\linewidth}  
	    \includegraphics[width=\linewidth]{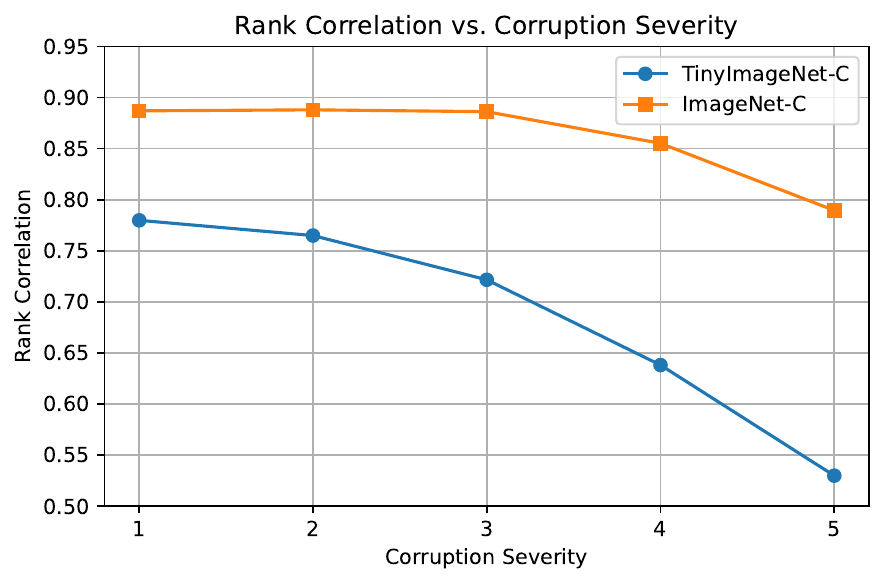}
        \caption{Rank correlation measured on ImageNet-C and TinyImageNet-C with different corruption degrees. }
	\label{fig:corruption}
	\end{subfigure}
	\caption{Analyses of pretraining strategies, model architectures, and degree of distribution shift.}
	\label{fig:ablation}
\end{figure*}

From the perspective of the evaluated algorithms, we can see that ATC, NI, Agreement, and COTT achieve relatively desirable overall results in performance prediction, even some of them are quite simple and straightforward compared with the rest of algorithms. Nevertheless, all of them still fail on many datasets, indicating that they are not universal enough to address diverse and complicated scenarios.  
It is also worth noting that the four algorithms adopt completely different practices: ATC employs model confidence, NI leverages invariance under data augmentations, Agreement takes advantage of the property of generalization disagreement equality~\citep{jiang2021assessing}, and COTT focuses on pseudo label shift. 
This implies that it is still not clear which direction is the correct and most promising one in the area of OOD performance prediction, and the development of performance prediction algorithms is far from convergence.

\subsection{Analysis of evaluation metrics}
\label{sec:metric}
We mainly focus on the comparison between Spearman's rank correlation $\rho$ and the coefficient of determination $R^2$. 
We pick three showcases for analyses shown in ~\Cref{fig:metric}: MaNo applied on Real domain of OfficeHome, MDE applied on Art domain of PACS, and Dispersion applied on Product domain of OfficeHome. 
In \Cref{fig:pattern1}, it shows that $R^2$ is relatively high but $\rho$ is quite low, and we can clearly see that there are several outliers with accuracies lower than $0.4$. 
When there are outliers, $R^2$ could become rather large without really exhibiting a strong linear correlation. 
In such a case, $\rho$ is more appropriate. 
In \Cref{fig:pattern2}, the scores and the ground truth accuracies are reversely correlated but $R^2$ is very high since there is a strong linear correlation. 
In such a case, $\rho$ is also more appropriate. 
\Cref{fig:pattern3} shows the opposite, i.e. $R^2$ is relatively low but $\rho$ is quite high. We can see that although the scores and the ground truth accuracies are not linearly correlated, they still show a relatively monotonic pattern, which also matches our requirement of evaluation. 
Overall, considering these failure cases of $R^2$, we treat Spearman's rank correlation $\rho$ as a more proper and effective evaluation metric for OOD performance prediction, and we adopt it in our main experiments. 
Discussion related to other metrics can be referred to in~\Cref{appendix:metric}.

\subsection{Analysis of OOD generalization methods}
In previous works, performance prediction algorithms were only applied on models trained with simple ERM. 
To investigate the effect of OOD generalization methods that are used to train models, we choose four representative OOD generalization methods: RSC~\citep{huang2020self}, Mixup~\citep{zhang2018mixup}, SWAD~\citep{cha2021swad}, and CORAL~\citep{sun2016deep}, and train models on PACS with them. 
Results are in~\Cref{tab:algorithm}. We can see that the change of OOD generalization methods has a large influence in some cases, e.g. MaNo applied to models trained with ERM and RSC, COTT applied to models trained with ERM and Mixup. 
Overall, when applied to models trained with OOD generalization methods, performance prediction algorithms could achieve higher rank correlations compared with models trained with ERM.
This could be because OOD generalization methods help reduce the performance gap between training and test distribution, which might be easier for performance prediction algorithms to work.

\subsection{Analysis of pretraining strategies}

We investigate pretraining strategies including supervised pretraining, MoCo (MoCo-v2 for ResNet-50 and MoCo-v3 for ViT-B/16), and CLIP on PACS. 
Results of ATC, NI, Agreement, and COTT, the four most effective algorithms, are shown in~\Cref{fig:pretrain}. We can see that pretraining strategies have a large influence on the effectiveness of performance prediction. 
For ATC, NI, and COTT, the rank correlation increases as the pretraining strategy changes from supervised to MoCo and CLIP. 
This could be because MoCo and CLIP are contrastive learning strategies that do not directly rely on the $y$ label, so models initialized with these pretrained weights are less likely to encounter over-confidence and could be better calibrated, making it easier for performance prediction algorithms to apply to. 
For Agreement, the supervised pretraining strategy performs best, which could be because the property of generalization disagreement equality~\citep{jiang2021assessing} holds better with supervised pretrained weight initialization.

\subsection{Analysis of model architectures}

To investigate the effect of model architectures on the results of performance prediction, we compare ResNet-50 and ViT-B/16 on PACS. 
From~\Cref{fig:arch}, we can see that for most algorithms, changing to a larger architecture results in even lower rank correlations except for ATC. 
This indicates that it is generally harder to predict performances of larger models, but confidence-based algorithms are worth being explored for scaling to larger models in future research. 

\subsection{Analysis of degree of distribution shift}

To characterize the degree of distribution shift, we investigate NI on TinyImageNet-C and ImageNet-C, both of which have 5 degrees of distribution shift depicted by the severity of corruption. 
\Cref{fig:corruption} shows that the effectiveness of NI decreases as the corruption severity grows. This could indicate that performance prediction algorithms work better with smaller distribution shifts. 
However, as mentioned in~\Cref{sec:mainexp} and revealed in~\Cref{tab:benchmark}, datasets like NICO++ and ObjectNet also exhibits strong distribution shift, but many algorithms show their effectiveness on these datasets, indicating the requirement of further in-depth analyses.

\subsection{Analysis of subpopulations}

In~\Cref{tab:benchmark}, results of our benchmark show that current performance prediction algorithms fail on subpopulation shift datasets where subpopulations are induced by attributes and categories with a strong spurious correlation. 
Here we conduct experiments by conducting performance prediction on each category, which also formulates subpopulations but without strong spurious correlations. 
We choose five algorithms that do not require diversity of category labels. 
From~\Cref{tab:category}, we can see that the phenomenon of failure on subpopulation shift datasets does not occur here. 
This implies that spurious correlations play a key role in the failure of performance predictions under subpopulation shift situations, which could be paid more attention to in the future. 

\begin{table}[htbp]
\centering
\caption{Rank correlation measured on each category of PACS. 
}
\label{tab:category}
\resizebox{0.95\columnwidth}{!}{%
\begin{tabular}{@{}c|ccccc|c@{}}
\toprule
Category & ATC   & DOC   & NI    & MaNo  & Agreement & \multicolumn{1}{c}{Avg.} \\ \midrule
$y=0$    & 0.737 & 0.782 & 0.822 & 0.425 & 0.915     & 0.736                       \\
$y=1$    & 0.817 & 0.811 & 0.847 & 0.328 & 0.895     & 0.740                       \\
$y=2$    & 0.737 & 0.769 & 0.779 & 0.445 & 0.892     & 0.725                       \\
$y=3$    & 0.815 & 0.752 & 0.760 & 0.360 & 0.916     & 0.721                       \\
$y=4$    & 0.770 & 0.671 & 0.790 & 0.277 & 0.677     & 0.637                       \\
$y=5$    & 0.878 & 0.829 & 0.831 & 0.511 & 0.880     & 0.786                       \\
$y=6$    & 0.798 & 0.798 & 0.851 & 0.409 & 0.838     & 0.739                       \\ \bottomrule
\end{tabular}%
}
\end{table}

%% file: sec/5_con.tex
\section{Conclusion}
\label{sec:con}
In this paper, we propose a large and comprehensive benchmark named \OURS. It includes \NUMDATA~OOD test datasets and \NUMALG~OOD performance prediction algorithms, covering diverse types of distribution shifts. 
It provides a testbench of \NUMMODEL~off-the-shelf trained models, which greatly reduces the burden of model training for future researchers and enables fair comparisons between different algorithms. 
It also provides a codebase so that a newly proposed algorithm can be easily added without complicated code implementation. 
The experimental results show that although current performance prediction algorithms exhibit effectiveness on certain types of shifts, e.g. synthetic corruptions, they are not universal enough to address all kinds of complex real-world shifts, which are left for future research. 

%% file: sec/X_suppl.tex
\clearpage
\setcounter{page}{1}
\maketitlesupplementary

\section{Appendix}

\subsection{Benchmark details}
\label{appendix:detail}

\subsubsection{Datasets}
The source datasets include ImageNet~\citep{deng2009imagenet}, CIFAR-10/100~\citep{krizhevsky2009learning}, and datasets adopted in WILDS~\citep{koh2021wilds}, domain generalization~\citep{gulrajani2020search, yu2024rethinking}, and subpopulation shift~\citep{yang2023change}. The OOD test datasets for each type of source dataset are as follows. The detailed downloading links can be found in our github repository.

\paragraph{CIFAR} OOD datasets as variants of CIFAR are listed below (all are variants of CIFAR-10 except CIFAR-100-C):
\begin{itemize}
    \item \textbf{CIFAR-10-C}~\citep{hendrycks2019benchmarking}: Add synthetic corruptions to the test dataset of CIFAR-10 with 5 degrees for each of all 20 corruptions.
    \item \textbf{CIFAR-100-C}~\citep{hendrycks2019benchmarking}: Add synthetic corruptions to the test dataset of CIFAR-100. 
    \item \textbf{CIFAR-10.1}~\citep{recht2018cifar}: A dataset whose collectors try their best to simulate the original data collection process of CIFAR-10. 
    \item \textbf{CIFAR-10.2}~\citep{lu2020harder}: Similar to CIFAR-10.1 but more difficult. 
    \item \textbf{CINIC-10}~\citep{darlow2018cinic}: An extension of CIFAR-10 by adding downsampled ImageNet images.
    \item \textbf{STL-10}~\citep{coates2011analysis}: A dataset collected differently from CIFAR-10 while sharing 9 common categories.  
\end{itemize}

\paragraph{ImageNet} OOD datasets as variants of ImageNet are listed below:
\begin{itemize}
    \item \textbf{ImageNet-C}~\citep{hendrycks2019benchmarking}: Add synthetic corruptions to validation data of ImageNet with 5 degrees for each of all 19 corruptions.
    \item \textbf{TinyImageNet-C}~\citep{hendrycks2019benchmarking}: A Smaller version of ImageNet-C with only 15 corruptions, 200 classes and size $64\times 64$. There are also 5 degrees for each corruption.
    \item \textbf{ImageNet-S}~\citep{wang2019learning}: ``S" stands for ``Sketch". It is a collection of sketches of all 1,000 classes of ImageNet.
    \item \textbf{ImageNet-R}~\citep{hendrycks2021many}: ``R" stands for ``Rendition". It includes 200 classes of ImageNet with various styles of images like art, cartoons, toys, video games, etc. 
    \item \textbf{ImageNet-V2}~\citep{recht2019imagenet}: A dataset whose collectors try their best to simulate the original data collection process of ImageNet. It includes 3 versions.
    \item \textbf{ImageNet-A}~\citep{hendrycks2021natural}: A dataset collected in a adversarial way, i.e. filtering the samples that were misclassified by the SOTA model at that time during data collection. It includes 200 classes of ImageNet. 
    \item \textbf{ImageNet-Vid}~\citep{shankar2019systematic}: Consist of many continuous video frames to create temporal shift, including 30 classes of ImageNet.
    \item \textbf{ObjectNet}~\citep{barbu2019objectnet}: A dataset collected by taking photos from all kinds of strange and uncommon views. It has 313 classes, 113 of which are shared by ImageNet and used for performance prediction.
\end{itemize}

\paragraph{WILDS~\citep{koh2021wilds}} We choose 6 datasets of WILDS to include in our benchmark. For each dataset, we follow the default setting of WILDS to divide the training, validation, and test set. More details of the datasets could also be referred to in WILDS~\footnote{https://github.com/p-lambda/wilds}.
\begin{itemize}
    \item \textbf{iWildCam}~\citep{beery2021iwildcam}: Photos of wildlife taken in different camera locations. 
    \item \textbf{FMoW}~\citep{christie2018functional}: Satellite images taken at different times and regions. 
    \item  \textbf{Camelyon17}~\citep{bandi2018detection}: Histopathological images collected from different hospitals. 
    \item \textbf{RxRx1}~\citep{sypetkowski2023rxrx1}: Cell images obtained by fluorescent microscopy under different experimental batches.
    \item \textbf{Amazon}~\citep{ni2019justifying}: Predict ratings from reviews written by different individuals. 
    \item \textbf{CivilComments}~\citep{borkan2019nuanced}: Comments (some are toxic) collected online with demographic metadata.
\end{itemize}

\paragraph{Domain generalization (DG)} We adopt all commonly used datasets in DG. For datasets of 4 domains, we follow~\citet{gulrajani2020search} to leave one domain out each time. For datasets of 6 domains, we follow~\citet{yu2024rethinking} to leave one group (two domains) out each time.
\begin{itemize}
    \item \textbf{PACS}~\citep{li2017deeper}: It has 7 categories and 4 domains of different styles: photo, art painting, carton, sketch. 
    \item \textbf{VLCS}~\citep{khosla2012undoing}: It is assembled by 5 common categories shared across 4 datasets: PASCAL, LABELME, CALTECH, SUN.
    \item \textbf{OfficeHome}~\citep{venkateswara2017deep}: It has 65 categories and 4 domains of different styles: Art, Product, Clipart, Real-World.
    \item \textbf{TerraInc}~\citep{beery2018recognition}: It has 10 categories and 4 domains with different camera locations. 
    \item \textbf{DomainNet}~\citep{peng2019moment}: It has 345 categories and 6 domains: clipart, quickdraw, infograph, real, painting, sketch. 
    \item \textbf{NICO++}~\citep{zhang2023nico++}: It has 60 categories and 6 public domains: autumn, dim, grass, outdoor, rock, water. 
\end{itemize}

\paragraph{Subpopulation shift} We adopt three datasets with the most standard subpopulation shift:
\begin{itemize}
    \item \textbf{Waterbirds}~\citep{sagawa2020distributionally}: A half-synthetic dataset of predicting the type of bird (waterbird or landbird). The spurious attribute is the background (water or land). In our benchmark, we treat blond hair male as the target worst group. We treat ``Waterbird-Land" as the target worst group. 
    \item \textbf{CelebA}~\citep{liu2015deep}: A large facial dataset with abundant attribute annotations. We follow~\citet{sagawa2020distributionally} by treating it as a binary classification task of predicting whether it is a face of blond hair or not. The spurious attribute is sex. In our benchmark, we treat blond hair male as the target worst group. 
    \item \textbf{CheXpert}~\citep{irvin2019chexpert}: A medical imaging dataset of predicting whether the patient is ill given the image of chest X-Ray. The spurious attributes are race and sex. We treat ill while female as the target worst group. 
\end{itemize}

\paragraph{Other remarks} For ObjectNet, although it has 313 categories, only 113 of them are shared with ImageNet. Thus we only use the 113 categories in our experiments. For STL-10, only 9 out of 10 categories are shared with CIFAR-10, so we only use 9 categories. 
For the \NUMDATA~datasets used in our benchmark, CivilComments and Amazon are text classification while all others are image classification. 
For these two text datasets, we do not run NI since it is not very straightforward to conduct data augmentation as implementations of neighborhood on text data. 

For ImageNet-C and Tiny-ImageNet-C both with 5 degrees of corruptions, in the benchmark leaderboard, i.e. \Cref{tab:benchmark}, we only adopt the 5th degree as OOD test data to reduce computational burdens. This also creates a more challenging setting compared with using the whole test dataset since \Cref{fig:corruption} reveals that current algorithms are less effective with the increase of corruption severity. 

For DG datasets, when we adopt the leave-one-domain-out or leave-one-group-out setting, we calculate rank correlations for each pair of source and target, and take the average of rank correlations as the final result.

Note that COT and COTT fail on subpopulation shift datasets like CheXpert, which could be because COT and COTT assume a small shift of $P(Y)$. 
Though it might be unfair to compare methods on what they are not specified for, we still compare them with other methods on subpopulation shift datasets in our benchmark to understand their capability boundaries and to ensure the completeness of our benchmark. 
For new methods, we do not require them to be SOTA on all datasets to demonstrate their value.

\subsubsection{Model architectures}
\paragraph{Networks for ImageNet} We directly use models provided in Torchvision 0.21.0\footnote{https://docs.pytorch.org/vision/0.21/models.html}.
Note that there are 115 off-the-shelf models, but 5 of them cannot be run on a 24GB GPU, including: EfficientNet-B7, RegNetY128-GF (two weights), ViT-H/14 (IMAGENET1K\_SWAG\_E2E\_V1), ViT-L/16 (IMAGENET1K\_SWAG\_E2E\_V1). Another model VGG16 (IMAGENET1K\_FEATURES) is also excluded since it does not have a classification head. Thus we only include 109 models.

\paragraph{Networks for CIFAR-10} We employ 20 model architectures from the public repository\footnote{https://github.com/kuangliu/pytorch-cifar}, which are listed below:
\begin{itemize}
    \item \textbf{DenseNet} with 121 and 161 layers~\citep{huang2017densely}.
    \item \textbf{Deep Layer Aggregation (DLA)}, standard DLA and its simplified variant SimpleDLA~\citep{yu2018deep}.
    \item \textbf{Dual Path Networks (DPN)}, specifically the DPN-92 variant~\citep{chen2017dual}.
    \item \textbf{EfficientNet}, specifically the B0 variant~\citep{tan2019efficientnet}.
    \item \textbf{GoogLeNet}, a member of the Inception family~\citep{szegedy2015going}.
    \item \textbf{MobileNet}, both V1 and V2~\citep{sandler2018mobilenetv2}.
    \item \textbf{RegNet} with configurations X\_200 and X\_400~\citep{radosavovic2020designing}.
    \item \textbf{ResNet} with 18, 50, and 101 layers~\citep{he2016deep,he2016identity}.
    \item \textbf{ResNeXT} with various widths and depths (2x64d, 32x4d)~\citep{xie2017aggregated}.
    \item \textbf{SENet}, specifically SENet-18 ~\citep{hu2018squeeze}.
    \item \textbf{ShuffleNet}, specifically V2 and G2~\citep{zhang2018shufflenet, ma2018shufflenet}.
    \item \textbf{VGG} with 19 layers~\citep{simonyan2014very}.
\end{itemize}

\paragraph{Networks for CIFAR-100} We employ 36 model architectures from the public repository\footnote{https://github.com/weiaicunzai/pytorch-cifar100}, which are listed below:
\begin{itemize}
    \item \textbf{DenseNet} with 121, 161, and 201 layers~\citep{huang2017densely}.
    \item \textbf{Inception} models: GoogLeNet, InceptionV4, InceptionV3, and InceptionResNetV2~\citep{szegedy2015going, szegedy2015rethinking,
    szegedy2016inceptionv4}.
    \item \textbf{MobileNet}, both V1 and V2~\citep{sandler2018mobilenetv2}.
    \item \textbf{NASNet}~\citep{zoph2018learning}.
    \item \textbf{ResNet} with 18, 34, 50, 101, and 152 layers~\citep{he2016deep,he2016identity}.
    \item \textbf{ResNeXT}, configurations 50\_32x4d, 101\_32x4d and 152\_32x4d~\citep{xie2017aggregated}.
    \item \textbf{SENet} with 18, 34, 50, 101, and 152 layers~\citep{hu2018squeeze}.
    \item \textbf{ShuffleNet}, both V1 and V2~\citep{zhang2018shufflenet,ma2018shufflenet}.    
    \item \textbf{SqueezeNet}, V1~\citep{i2016squeezenet}.
    \item \textbf{StochasticDepth} with 18, 34, 50 and 101 layers~\citep{huang2016deep}
    \item \textbf{VGG} with 11, 13, 16, and 19 layers~\citep{simonyan2014very}.
    \item \textbf{WideResNet}~\citep{zagoruyko2016wide}.
    \item \textbf{Xception}~\citep{chollet2017xception}.
\end{itemize}

For networks used for WILDS, we directly follow settings provided by \citet{koh2021wilds}. 
For networks used for datasets from domain generalization and subpopulation shifts, we have stated in \Cref{sec:training}.

\subsection{More experimental results and analyses}

\subsubsection{Evaluation metrics}
\label{appendix:metric}

We conduct explorations of other possible evaluation metrics. 
For datasets like ImageNet-A, most models might perform poorly. In such a case, we do not really care about how the poorly performing models can be ranked, but care about the best performing models. 
Thus it could be more reasonable to calculate metrics using only, for example, top 10\% models.
In \Cref{appendix-fig:metric}, we plot a histogram of real performances on ImageNet-A, and scatter plots of real performances and predicted scores of MaNo on ImageNet-A and ATC on NICO++ (water and outdoor as target domains) as showcases. 
To reduce influence of poorly performing models, we design two metrics. (1) Precision$@10\%$: Proportion of models predicted as top 10\% whose real performance also ranks top 10\%; (2) $\rho@10\%$: rank correlation among predicted top 10\% models. Note that when there are fewer than 100 models, we calculate top 10 instead of 10\%.
\Cref{fig:hist} shows most models perform poorly on ImageNet-A and \Cref{fig:mano-imageneta} shows that MaNo fails to rank most models (blue points) correctly, achieving a low rank correlation in \Cref{tab:metric}, 
while MaNo obtains high values of the two new metrics in \Cref{tab:metric} since it ranks the predicted top 10\% models (red points) well. 
Meanwhile, for ATC on NICO++, a case with few bad models, rank correlation could be effective enough for evaluation. 
Thus the two metrics are effective as complementary metrics to rank correlation. We suggest that these metrics could also be calculated on certain datasets in future research.

\begin{table}[h]
\centering
\caption{Comparing different metrics.}
\label{tab:metric}
\resizebox{0.3\textwidth}{!}{%
\begin{tabular}{@{}c|ccc@{}}
\toprule
Dataset             & $\rho$ & Precision@$10\%$ & $\rho@10\%$ \\ \midrule
ImageNet-A & 0.417  & 0.818                  & 0.927       \\ \midrule
NICO++      & 0.952  & 0.800                  & 0.455       \\ \bottomrule
\end{tabular}%
}
\end{table}

\begin{figure}[h]
	\centering
        \begin{subfigure}{0.3\linewidth}  
	    \includegraphics[width=\linewidth]{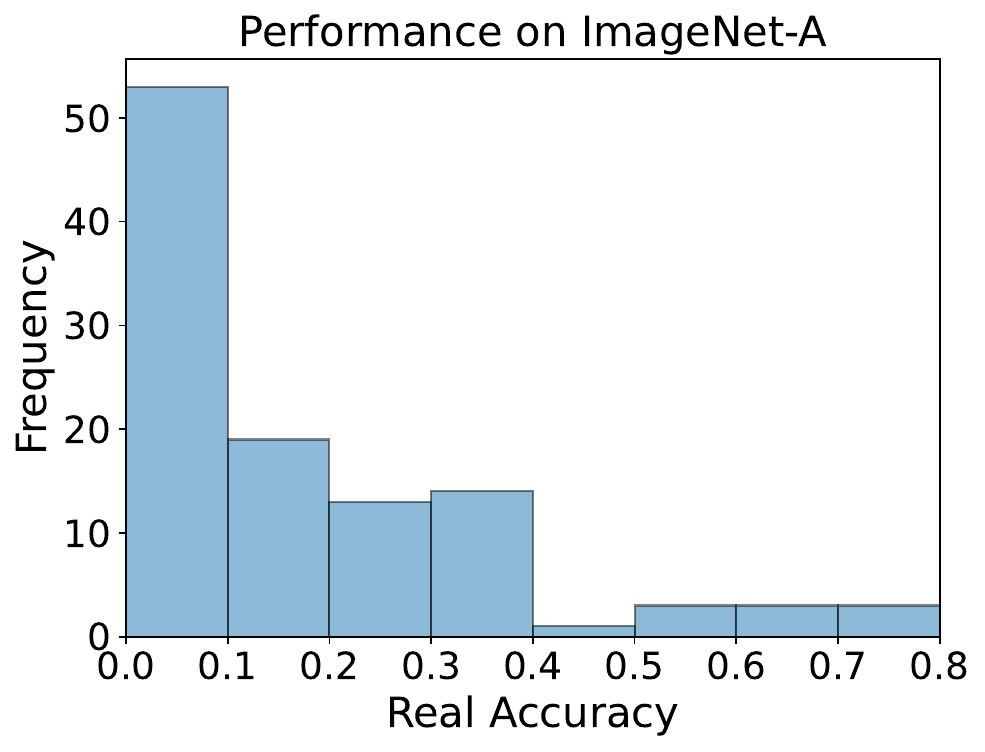}
            \caption{Histogram of accuracies for ImageNet-A.}
	    \label{fig:hist}
	\end{subfigure}
	\begin{subfigure}{0.3\linewidth}  
	    \includegraphics[width=\linewidth]{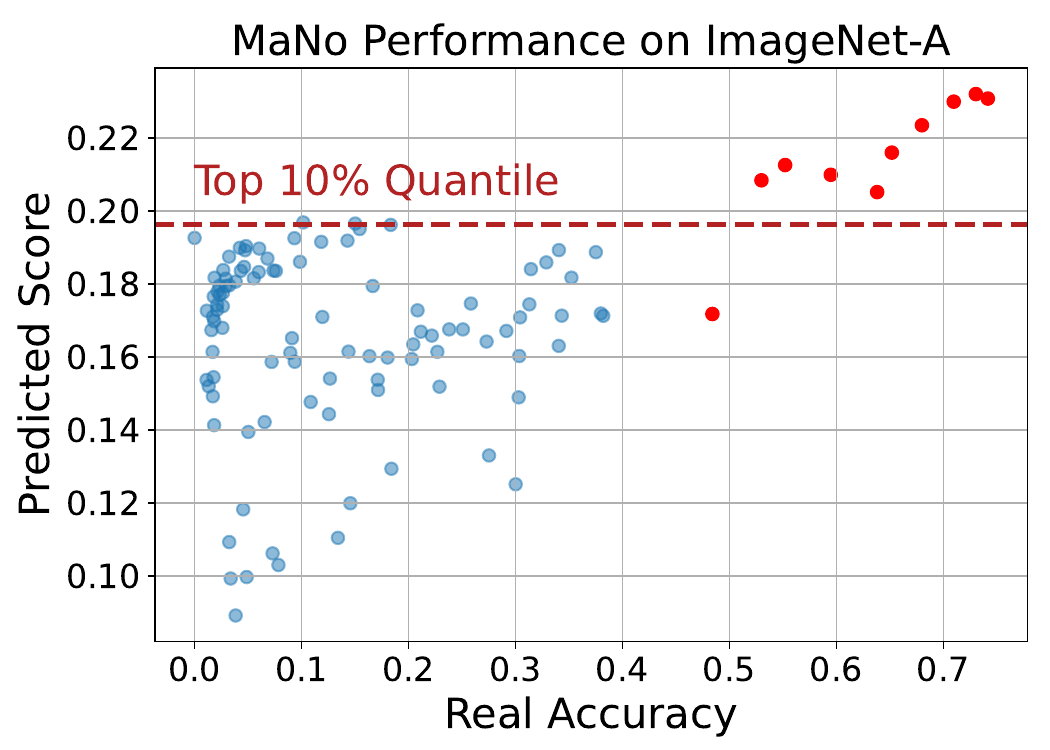}
            \caption{MaNo on ImageNet-A.}
	    \label{fig:mano-imageneta}
	\end{subfigure}
        \begin{subfigure}{0.3\linewidth}  
	    \includegraphics[width=\linewidth]{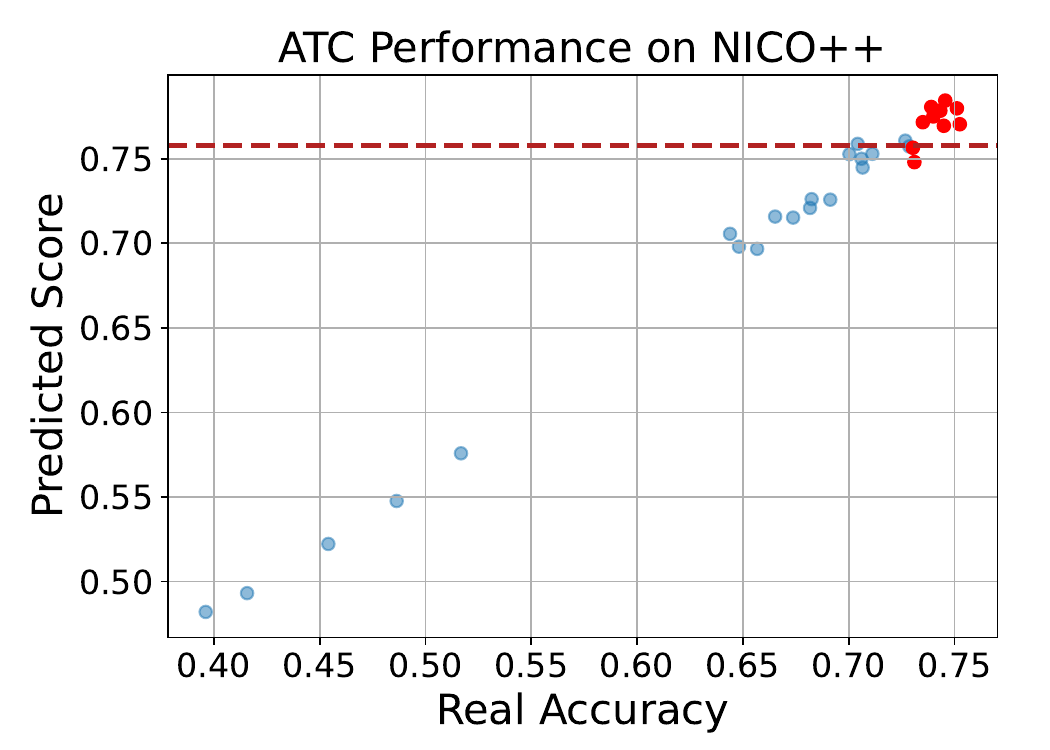}
            \caption{ATC on NICO++.}
	    \label{fig:atc-nico}
	\end{subfigure}
	\caption{Distribution of real performances and predicted scores.}
	\label{appendix-fig:metric}
\end{figure}

\subsection{Agreement and calibration}
\begin{figure}[h]
	\centering
            \includegraphics[width=\linewidth]{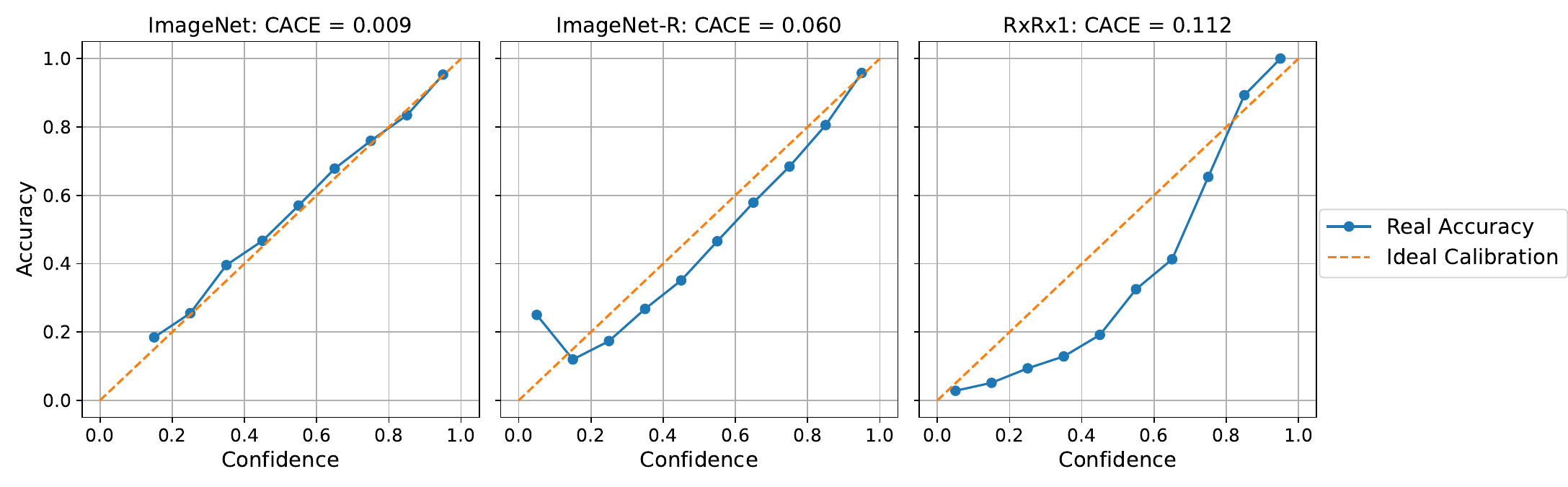}
	\caption{Calibration on ImageNet, ImageNet-R, and RxRx1.}
	\label{fig:calib}
\end{figure}

In the original paper of agreement~\citep{jiang2022assessing}, it requires class-wise calibration for the ensemble of models to satisfy the Generalization Disagreement Equality (GDE). 
We draw plots of model accuracy against confidence in \Cref{fig:calib} for ID (ImageNet) and OOD (ImageNet-R, RxRx1) cases and calculate Class Aggregated Calibration Error (CACE). 
The assumption of calibration does not hold on two OOD scenarios where CACE greatly increases and the curves deviate from the ideal line.
However, Agreement gets a higher rank correlation on RxRx1 (0.979) than ImageNet-R (0.926) while CACE of the former is much higher. 
It indicates that the effectiveness of agreement in performance prediction does not necessarily rely on calibration.

